%% file: main.tex
\documentclass[runningheads]{llncs}
\usepackage[utf8]{inputenc}
\usepackage[T1]{fontenc}
\usepackage{booktabs}
\usepackage{graphicx}
\usepackage{xspace}
\usepackage[colorlinks=true,allcolors=blue]{hyperref}
\usepackage{ifthen}
\usepackage[acronym]{glossaries}
\glsdisablehyper
\usepackage[dvipsnames,table]{xcolor}
\usepackage[capitalise]{cleveref} %
\usepackage[sort,compress]{cite}
\usepackage[caption=false,farskip=0pt]{subfig}
\usepackage{multirow}
\usepackage{svg}

\newif\ifarxiv
\arxivtrue

\newcommand\customorcidAuthor[1]{\hspace*{-1mm}
\href{https://orcid.org/#1}{\includegraphics[width=0.3cm]{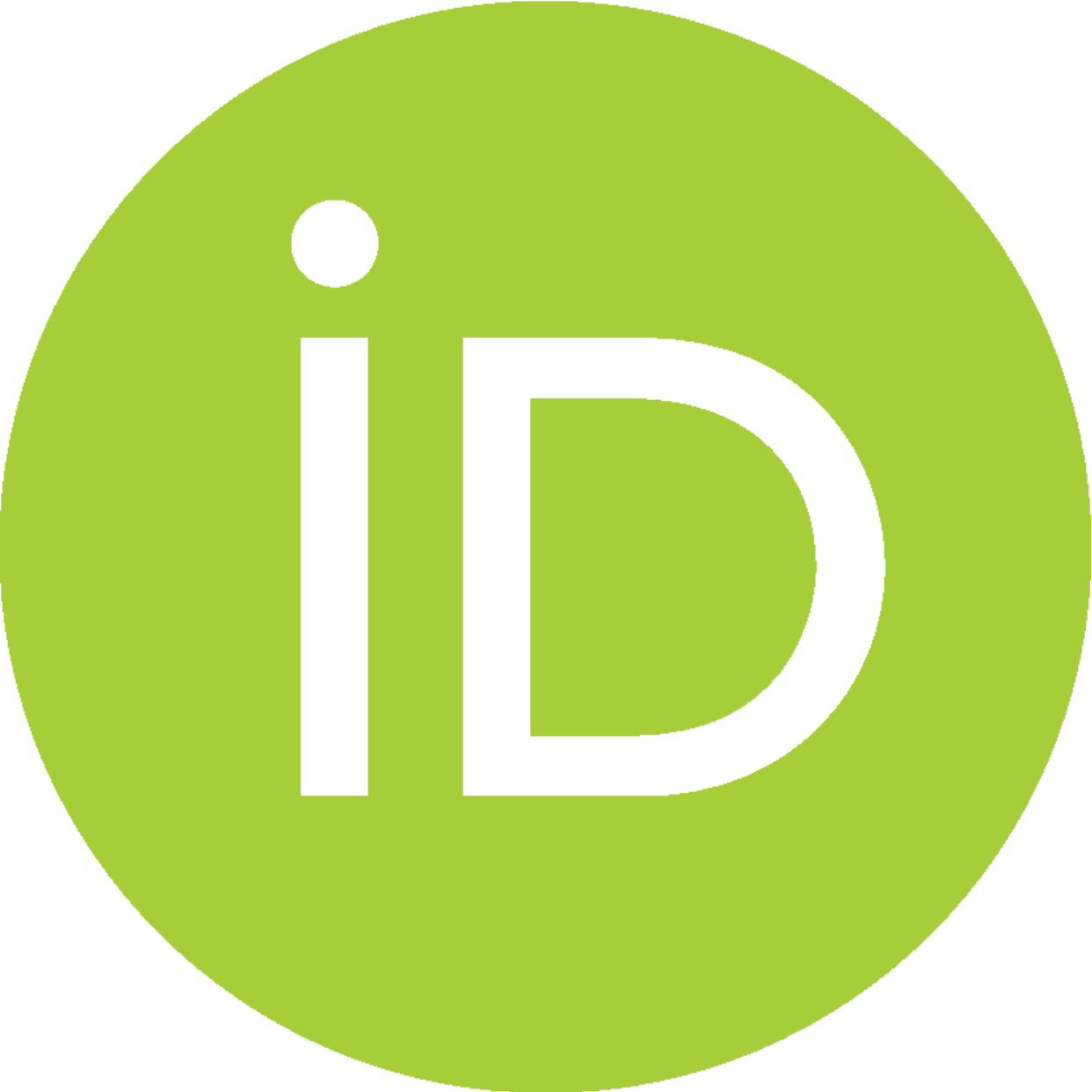}}\hspace*{-1mm}
} %

\newcommand*{\RL}[2][]{\textcolor{Rhodamine}{[\textbf{\ifthenelse{\equal{#1}{}}{RL}{RL(#1)}}: #2]}}

\newcommand*{\VN}[2][]{\textcolor{Orange}{[\textbf{\ifthenelse{\equal{#1}{}}{VN}{VN(#1)}}: #2]}}

\newacronymstyle{long-short-br}
{%
  \GlsUseAcrEntryDispStyle{long-short}%
}%
{%
  \GlsUseAcrStyleDefs{long-short}%
}
\setacronymstyle{long-short-br}

\usepackage{transparent}
\usepackage{tikz}
\ifarxiv
    \newcommand\copyrighttext{%
      \scriptsize This is an author-prepared version. The final published version is available from \textit{Springer} (DOI: \href{https://doi.org/10.1007/978-3-032-31936-4_18}{\textcolor{blue}{https://doi.org/10.1007/978-3-032-31936-4\_18}}).}
    \newcommand\copyrightnotice{%
    \vspace{-5mm}
    \begin{tikzpicture}[remember picture,overlay]
    \node[anchor=south,yshift=80pt,xshift=0pt] at (current page.south) {\fbox{\transparent{0.85}\parbox{\dimexpr0.74\textwidth-\fboxsep-\fboxrule\relax}{\copyrighttext}}};
    \end{tikzpicture}%
    }
\else
\fi

\begin{document}
\title{ICPR 2026 Competition on\\Low-Resolution License Plate Recognition}
\author{Rayson~Laroca\inst{1,2,*}\customorcidAuthor{0000-0003-1943-2711} \and
Valfride~Nascimento\inst{1}\customorcidAuthor{0000-0002-7416-613X} \and
Donggun~Kim\inst{3}\customorcidAuthor{0009-0007-4745-9731} \and 
Sanghyeok~Chung\inst{3}\customorcidAuthor{0009-0002-2257-4729} \and
Subin~Bae\inst{3}\customorcidAuthor{0009-0002-0365-6155} \and
Uihwan~Seo\inst{3}\customorcidAuthor{0009-0007-5457-8720} \and
Seungsang~Oh\inst{3}\customorcidAuthor{0000-0003-4975-9977} \and
Chi~M.~Phung\inst{4,6}\customorcidAuthor{0009-0009-2514-9729} \and
Minh~G.~Vo\inst{5,6}\customorcidAuthor{0009-0000-8890-0449} \and
Xingsong~Ye\inst{7,8}\customorcidAuthor{0009-0007-4046-7229} \and
Yongkun~Du\inst{7,8}\customorcidAuthor{0009-0005-3114-2188} \and
Yuchen~Su\inst{7,8}\customorcidAuthor{0000-0001-7743-8260} \and
Zhineng~Chen\inst{7,8}\customorcidAuthor{0000-0003-1543-6889} \and
Sunhee~Heo\inst{9}\customorcidAuthor{0009-0000-3495-9138} \and 
Hyangwoo~Lee\inst{9}\customorcidAuthor{0009-0005-5110-0751} \and
Kihyun~Na\inst{9}\customorcidAuthor{0009-0001-3827-5371} \and
Khanh~V.~Vu~Nguyen\inst{4,6}\customorcidAuthor{0009-0001-1075-213X} \and
Sang~T.~Pham\inst{4,6}\customorcidAuthor{0009-0004-5075-6241} \and
Duc~N.~N.~Phung\inst{4,6}\customorcidAuthor{0009-0001-0997-3971} \and
Trong~P.~Le\inst{4,6}\customorcidAuthor{0009-0003-3072-363X}\hspace{0.05mm} \and Vy~N.~Vo~Tran\inst{4,6}\customorcidAuthor{0009-0004-0728-1885}\hspace{0.05mm} \and
David~Menotti\inst{1}\customorcidAuthor{0000-0003-2430-2030}
}
\authorrunning{R. Laroca et al.}
\institute{
Federal University of Paraná, Brazil \and
Pontifical Catholic University of Paraná, Brazil \and 
Korea University, Republic of Korea \and %
University of Information Technology, Vietnam \and %
Ho Chi Minh University of Technology, Vietnam \and %
Vietnam National University, Vietnam \and %
Fudan University, China \and %
Shanghai Key Laboratory of Multimodal Embodied AI, China \and %
Handong Global University, Republic of Korea \\[1.75ex]
* \email{rayson@ppgia.pucpr.br}
}
\maketitle              %

\input{0-acronyms}
\input{0-variables}

\ifarxiv
    \copyrightnotice
\else
\fi

\input{0-abstract_keywords}

\input{1-introduction}
\input{2-competition}

\input{3-results}
\input{4-discussion}
\input{5-conclusions}

\input{0-acknowledgments}

\bibliographystyle{splncs04}
\bibliography{bibtex}

\end{document}

%% file: 0-acronyms.tex
\newacronym{alpr}{ALPR}{Automatic License Plate Recognition}
\newacronym{ema}{EMA}{Exponential Moving Average}
\newacronym{hr}{HR}{high-resolution}
\newacronym{ctc}{CTC}{Connectionist Temporal Classification}
\newacronym{lp}{LP}{License Plate}
\newacronym{lpd}{LPD}{License Plate Detection}
\newacronym{lpr}{LPR}{License Plate Recognition}
\newacronym{lr}{LR}{low-resolution}
\newacronym{lrlpr}{LRLPR}{Low-Resolution License Plate Recognition}
\newacronym{icpr}{ICPR}{International Conference on Pattern Recognition}
\newacronym{ocr}{OCR}{Optical Character Recognition}
\newacronym{ohem}{OHEM}{Online Hard Example Mining}
\newacronym{stn}{STN}{Spatial Transformer Network}

%% file: 0-variables.tex
\newcommand{\openalprbr}{OpenALPR\nobreakdash-BR\xspace}
\newcommand{\rodosol}{RodoSol\nobreakdash-ALPR\xspace}
\newcommand{\rodosolalpr}{\rodosol}
\newcommand{\ufpralpr}{UFPR\nobreakdash-ALPR\xspace}
\newcommand{\ufprsrplates}{UFPR\nobreakdash-SR\nobreakdash-Plates\xspace}

\newcommand{\dataset}{LRLPR\nobreakdash-26\xspace}
\newcommand{\supplementary}{\url{https://raysonlaroca.github.io/supp/lrlpr26/}}
\newcommand{\competitionURL}{\url{https://icpr26lrlpr.github.io/}}
\newcommand{\competitionCodabenchURL}{\url{https://www.codabench.org/competitions/12259/}}

\newcommand{\nTopRanked}{5\xspace}
\newcommand{\numParticipantsCodabench}{525\xspace}
\newcommand{\numTeams}{269\xspace}
\newcommand{\numCountries}{41\xspace}
\newcommand{\numSubmissionsTotalPublic}{1{,}164\xspace}
\newcommand{\numSubmissionsTeamsPublic}{118\xspace}
\newcommand{\numSubmissionsTotalBlind}{198\xspace}
\newcommand{\numSubmissionsTeamsBlind}{99\xspace}

%% file: 0-abstract_keywords.tex
\begin{abstract}

\gls*{lrlpr} remains a challenging problem in real-world surveillance scenarios, where long capture distances, compression artifacts, and adverse imaging conditions can severely degrade license plate legibility.
To promote progress in this area, we organized the ICPR 2026 Competition on Low-Resolution License Plate Recognition, the first competition specifically dedicated to \gls*{lrlpr} using real low-quality data collected under operationally relevant conditions.
The competition was based on the \dataset dataset, which comprises 20,000 training tracks and 3,000 test tracks; each training track contains five low-resolution and five high-resolution images of the same license plate.
Notably, a total of \numTeams teams from \numCountries countries registered for the competition, and \numSubmissionsTeamsBlind teams submitted valid entries in the Blind Test Phase.
The winning team achieved a Recognition Rate of 82.13\%, and four teams surpassed the 80\% mark, highlighting both the high level of competition at the top of the leaderboard and the continued difficulty of the task.
In addition to presenting the competition design, evaluation protocol, and main results, this paper summarizes the methods adopted by the top-5 teams and discusses current trends and promising directions for future research on \gls*{lrlpr}.
The competition webpage is available at \competitionURL

\keywords{Automatic license plate recognition \and Low-resolution \and Real-world scenarios \and Forensics \and Challenge.}
\end{abstract}

%% file: 1-introduction.tex
\section{Introduction}
\label{sec:introduction}

\setcounter{footnote}{0}
\glsresetall

\gls*{alpr} systems rely on image processing and pattern recognition techniques to detect and recognize \glspl*{lp} in images or videos. 
They are widely used in traffic law enforcement, electronic toll collection, and vehicle access control in restricted areas~\cite{weihong2020research,ismail2025automatic}.

With the evolution of general-purpose object detectors, particularly YOLO and its variants~\cite{terven2023comprehensive}, \gls*{lp} detection performance has approached saturation under standard imaging conditions~\cite{laroca2021efficient,silva2022flexible,ke2023ultra}. 
Recent studies have also reported very high recognition rates for complete \gls*{alpr} pipelines~\cite{laroca2023leveraging,rao2024license,laroca2025advancing}. 
However, such results are largely obtained from high-quality images, where \gls*{lp} characters are sharp, well-defined, and minimally affected by noise or compression artifacts.

In real-world surveillance environments, \gls*{lp} images are frequently captured at low resolution due to hardware constraints or the large distance between vehicles and cameras~\cite{nascimento2022combining,pan2024lpsrgan,gong2025lpdiff}. 
In addition, storage and bandwidth limitations often require strong compression, further degrading visual quality~\cite{schirrmacher2023benchmarking,nascimento2025toward,wojcik2026lplcv2}. 
As a result, characters may appear blurred, distorted, or barely distinguishable from the background, significantly increasing recognition difficulty~\cite{goncalves2019multitask,gong2024dataset,na2025mflpr2}.

Despite its clear practical importance, \gls*{lrlpr} remains a challenging and relatively underexplored problem with strong forensic and societal relevance~\cite{moussa2022forensic,schirrmacher2023benchmarking}.
Even state-of-the-art approaches struggle to exceed 50--60\% recognition accuracy on real-world low-quality images~\cite{nascimento2024enhancing,gong2025lpdiff,nascimento2025toward}. 
Enhancing recognition performance under such adverse conditions can significantly reduce investigative search spaces and support faster, more reliable decision-making in law enforcement workflows~\cite{maier2022reliability,schirrmacher2023benchmarking}.

Although several studies report promising results on \gls*{lrlpr}~\cite{moussa2022forensic,nascimento2022combining,kim2024afanet}, most rely on synthetically degraded images derived from high-resolution samples, typically using bicubic downsampling. Such simplified degradation models fail to capture the complex artifacts and variability present in real operational scenarios~\cite{gong2024dataset,gong2025lpdiff,na2025mflpr2}.
This limitation underscores the importance of benchmarks built from genuinely low-quality data collected under real-world conditions.

To foster progress in this direction, we organized the first \textit{Competition on Low-Resolution License Plate Recognition}, held in conjunction with the 2026 \gls*{icpr}\footnote{\competitionURL}.
The competition was based on the \dataset dataset, an expanded version of our previously released benchmark~\cite{nascimento2025toward}.
The dataset comprises $200{,}000$ images organized into $20{,}000$ training tracks, each containing five \gls*{lr} images and five \gls*{hr} images of the same \gls*{lp}, as well as $15{,}000$ images organized into $3{,}000$ test tracks, each containing five \gls*{lr} images of the same \gls*{lp}.
To the best of our knowledge, \dataset is the largest public dataset featuring real low- and high-resolution \glspl*{lp} acquired from the same~vehicles.

Remarkably, the competition attracted \numTeams\ teams from \numCountries\ countries worldwide, which submitted \numSubmissionsTeamsBlind\ entries during the final evaluation phase alone.
In this paper, we present an overview of the competition, including the dataset design, the evaluation protocol, the results, and a description of the five top-ranked approaches.
Our goal is to foster further advances in \gls*{lrlpr} and to contribute to the broader development of scene text recognition systems that can operate reliably under adverse imaging~conditions.

The rest of this paper is organized as follows. 
\cref{sec:competition} presents the competition, including the dataset, competition phases, evaluation protocol, participation statistics, and fairness rules. 
\cref{sec:results} reports the competition results and summarizes the top-ranked approaches. 
\cref{sec:discussion} discusses the main findings and implications of the competition. 
Finally, \cref{sec:conclusions} concludes the paper.

%% file: 2-competition.tex
\section{The 2026 LRLPR Competition}
\label{sec:competition}

\glsreset{lr}
\glsreset{hr}

The competition was designed with two main objectives.
First, it aimed to provide a broad view of recent advances and emerging trends in \gls*{lrlpr}. 
Second, it sought to bring together researchers from different institutions and backgrounds, fostering exchange and potential collaboration within the research~community.

The remainder of this section is organized as follows. \cref{sec:competition:training_data,sec:competition:test_data} describe the training and test data provided to participants.
\cref{sec:competition:dataset:privacy} discusses privacy considerations associated with the dataset.
\cref{sec:competition:phases-format} presents the competition phases and submission format, and \cref{sec:competition:evaluation_metrics} details the evaluation protocol.
\cref{sec:competition:participants} summarizes participation statistics, while \cref{sec:competition:fairness} outlines the main rules adopted to ensure a fair and transparent competition.

\subsection{Training Data}
\label{sec:competition:training_data}

The training data comprises 20,000 tracks, each containing five consecutive \gls*{lr} images and five consecutive \gls*{hr} images of the same \gls*{lp}, for a total of 200,000 images.
\cref{fig:tracks-extracted} shows all \gls*{lp} images from four example tracks.
The inclusion of \gls*{hr} images in the training set was intended to encourage participants to explore image enhancement strategies, such as super-resolution, that could improve recognition performance.

\input{2-figure-tracks}

The original videos (before \gls*{lp} cropping) were acquired with a rolling-shutter camera installed at the Federal University of Paraná, in Curitiba, Brazil, under conditions designed to resemble real-world surveillance scenarios.
Half of the training set, corresponding to 10,000 tracks, comes from the recently published \ufprsrplates dataset~\cite{nascimento2025toward}.
These tracks, hereinafter referred to as Scenario~A, were collected under relatively controlled conditions, such as daylight and no rain.
Although additional details about this dataset are available in~\cite{nascimento2025toward}, its acquisition procedure was essentially the same as that adopted for the remaining 10,000 tracks, described next as Scenario~B.

Scenario~B was collected specifically for this competition.
The same camera used in Scenario~A was employed, but it was oriented in a different direction, and the resulting data cover a broader range of environmental conditions, including rain and nighttime.
\cref{fig:dataset-samples} shows representative images used to build the dataset~(i.e., prior to \gls*{lp} detection and extraction), highlighting the diversity of vehicle categories and the variety of environmental~conditions.

\input{2-figure-samples-original}

The cropped \glspl*{lp} were obtained from video sequences of vehicles entering and leaving the roadway in opposite directions, as illustrated in \cref{fig:pipeline}.
All videos were recorded at a resolution of $1920\times1080$ pixels.
To detect and track \glspl*{lp}, we employed YOLOv11~\cite{yolov11}, motivated by the strong performance of the YOLO family in unconstrained scenarios~\cite{diwan2023object,laroca2025improving,lima2026toward}.
The detector was fine-tuned on widely used datasets in the \gls*{alpr} literature, including RodoSol-ALPR~\cite{laroca2022cross} and UFPR-ALPR~\cite{laroca2018robust}.
Detected \glspl*{lp} were then associated across frames using BoT-SORT~\cite{aharon2022botsort}.
From each track, the five patches extracted from the frames farthest from the camera were selected as the \gls*{lr} samples, whereas the five patches extracted from the nearest frames were selected as the \gls*{hr} samples.
To annotate the \gls*{lp} text, we applied the multi-task \gls*{ocr} model proposed by Gonçalves et al.~\cite{goncalves2018realtime} to each of the five \gls*{hr} images in a track and obtained the final label through majority voting~\cite{laroca2023leveraging}.

\input{2-figure-pipeline}

The dataset contains two \gls*{lp} layouts, namely Brazilian and Mercosur.
Brazilian \glspl*{lp} follow the pattern of three letters followed by four digits, whereas Mercosur \glspl*{lp} in Brazil follow the pattern of three letters, one digit, one letter, and two digits~\cite{laroca2022cross}.
The distribution of tracks for each layout across Scenarios~A and~B is detailed in \cref{tab:training_data}.
The higher concentration of Mercosur \glspl*{lp} in Scenario~B reflects their current prevalence in the Brazilian automotive fleet.
Furthermore, Scenario~B was designed with a strictly unique mapping---one track per \gls*{lp} (and thus per vehicle)---thereby increasing dataset diversity compared to Scenario~A.

\input{2-table-training-data}

Each track in Scenario~A is annotated with the \gls*{lp} text, the \gls*{lp} layout, and the~$(x,y)$ coordinates of the four \gls*{lp} corners~\cite{nascimento2025toward}.
In Scenario~B, however, corner annotations were not provided due to the high cost of manual labeling.
Although much of the acquisition pipeline was automated, all annotations were manually reviewed to ensure the reliability of the competition data.
During this process, occasional errors, mainly introduced by the \gls*{ocr} model, were identified and~corrected.

Participants were free to determine how much of the training data to reserve for validation.
As discussed in \cref{sec:results:top_teams}, the top-ranked teams adopted different validation strategies.
They were also free to choose how to combine predictions from the five \gls*{lr} images, for example, through majority voting, confidence-based selection, or temporal~modeling.

\subsection{Test Data}
\label{sec:competition:test_data}

The test set comprises 3,000 tracks, all collected exclusively from Scenario~B, with each track corresponding to a unique vehicle.
None of the \glspl*{lp} in the test set appears in the training data.
Its layout distribution matches that of the training set, comprising 600 tracks with Brazilian \glspl*{lp} and 2,400 with Mercosur~\glspl*{lp}.

The test set does not include annotations or \gls*{hr} images, as participants were required to predict the \gls*{lp} text using only the provided \gls*{lr}~images.

\subsection{Privacy Concerns}
\label{sec:competition:dataset:privacy}

We remark that \glspl*{lp} of vehicles registered in Brazil are not publicly linked to the personal information of their owners, which substantially reduces privacy risks. 
An \gls*{lp} uniquely identifies the vehicle itself and does not, by design, disclose personal data. 
Accordingly, the \gls*{lp} remains associated with the vehicle even after a transfer of ownership, underscoring its function as a vehicle identifier rather than a personal identifier. 
This characteristic has been previously discussed in prior studies introducing datasets collected in Brazil~\cite{oliveira2021vehicle,laroca2022cross,nascimento2025toward,lima2026toward}.

\subsection{Competition Phases and Submission Format}
\label{sec:competition:phases-format}

Although the competition had a single track, it was conducted in two phases.

The first phase, referred to as the \textit{Public Test Phase}, lasted approximately one month.
During this stage, participants had access to a subset of 1,000 out of the 3,000 test tracks, of which 200 corresponded to Brazilian \glspl*{lp} and 800 to Mercosur \glspl*{lp}.
This phase allowed participants to obtain feedback on their methods through Codabench~\cite{codabench} using part of the test set, while a public leaderboard was maintained on the same platform.
Each team was allowed up to five submissions per day, with a maximum of 25 submissions~overall.

Immediately after the end of the \textit{Public Test Phase}, the competition moved to its final stage, called the \textit{Blind Test Phase}, which lasted approximately one week.
At that point, the full test set, comprising the 3,000 tracks described in \cref{sec:competition:test_data}, was made available.
During this phase, the leaderboard became private: each team could view only its own score, but not the scores of the other teams.
To reduce leaderboard probing, each team was allowed up to three submissions in total, which also provided some margin in case of platform-related~issues.

In both phases, participants were required to submit a \texttt{.txt} prediction file in which each line consisted of \texttt{track\_id,plate\_text;confidence}, as illustrated below:
\begin{verbatim}
track_00001,ABC1234;0.9876
track_00002,DEF5678;0.6789
track_00003,GHI9012;0.4521
...
\end{verbatim}

The predicted \gls*{lp} strings and their corresponding confidence scores were used to calculate the evaluation metrics detailed in the following~section.

\subsection{Evaluation Protocol}
\label{sec:competition:evaluation_metrics}

The official ranking was determined by the \textit{Recognition Rate} computed on the Blind Test Set.
This metric evaluates performance at the track level and requires an exact match between the predicted and ground-truth \gls*{lp} strings~\cite{silva2022flexible,laroca2023do,wei2024efficient}.

Formally, the Recognition Rate is defined as:
\begin{equation}
\text{Recognition Rate} =
\frac{\text{Number of Correctly Recognized Tracks}}
{\text{Number of Tracks in the Test Set}}\;.
\end{equation}

When two or more teams achieved the same Recognition Rate, ties were broken using a secondary criterion, \textit{Confidence Gap}, which measures the model's ability to distinguish its correct from incorrect predictions based on its confidence scores. It is defined as the difference between the mean confidence assigned to correct predictions and that assigned to incorrect predictions.
A higher Confidence Gap indicates better-calibrated confidence~estimates.

The results reported in this paper are primarily based on the submissions made to the Codabench server\footnote{\competitionCodabenchURL}.
A script for computing these metrics is available at {\small\supplementary}.

\subsection{Participants}
\label{sec:competition:participants}

The competition attracted \numTeams teams from \numCountries countries, with the largest contingents coming from Vietnam, India, Mainland China, Taiwan, and the Republic of Korea. 
Among these, \numSubmissionsTeamsPublic teams submitted valid entries in the Public Test Phase, while \numSubmissionsTeamsBlind teams participated in the Blind Test Phase. This level of engagement, specifically acknowledged in official communications from the Codabench~team, is particularly remarkable for a competition without monetary prizes.

The competition would certainly have attracted even more teams if registration had been open to all interested participants.
However, registration and, consequently, access to the dataset were restricted to teams affiliated with accredited universities or research institutions.
This follows a common access policy adopted by several datasets in the \gls*{alpr} literature~\cite{hsu2013application,laroca2018robust,laroca2022cross}, including \ufprsrplates~\cite{nascimento2025toward}, which was incorporated into the official training data.

\subsection{Fairness}
\label{sec:competition:fairness}

To ensure a fair and transparent competition, a set of rules was established, following principles similar to those adopted in other recent competitions~\cite{coluccia2025drone,chen2025ntire_sr}.
Among the most important were the following.
Manual inspection or annotation of the test set was strictly prohibited.
At the same time, participants were allowed to train their methods using additional datasets, such as UFPR-ALPR~\cite{laroca2018robust} and RodoSol-ALPR~\cite{laroca2022cross}.
These rules were designed to discourage test-set leakage while still allowing methodological~flexibility.

%% file: 2-figure-tracks.tex
\begin{figure}[!hbt]
    \captionsetup[subfigure]{labelformat=empty,position=bottom,captionskip=0.75pt,justification=centering} 
    
    \centering

    \noindent
    \makebox[0.5\linewidth][c]{\scriptsize \phantom{i}LR Images}%
    \hfill
    \makebox[0.5\linewidth][c]{\scriptsize \phantom{i}HR Images}

    \vspace{0.3mm}

    \resizebox{0.997\linewidth}{!}{
    \subfloat[]{
    \includegraphics[width=0.15\textwidth, height=0.05\textwidth]{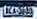}
    }\hspace{-2.1mm}
    \subfloat[]{
    \includegraphics[width=0.15\textwidth, height=0.05\textwidth]{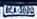}
    }\hspace{-2.1mm}
    \subfloat[]{
    \includegraphics[width=0.15\textwidth, height=0.05\textwidth]{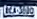}
    }\hspace{-2.1mm}
    \subfloat[]{
    \includegraphics[width=0.15\textwidth, height=0.05\textwidth]{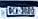}
    }\hspace{-2.1mm}
    \subfloat[]{
    \includegraphics[width=0.15\textwidth, height=0.05\textwidth]{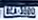}
    }\hspace{0.75mm}
    \subfloat[]{
    \includegraphics[width=0.15\textwidth, height=0.05\textwidth]{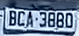}
    }\hspace{-2.1mm}
    \subfloat[]{
    \includegraphics[width=0.15\textwidth, height=0.05\textwidth]{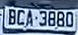}
    }\hspace{-2.1mm}
    \subfloat[]{
    \includegraphics[width=0.15\textwidth, height=0.05\textwidth]{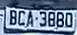}
    }\hspace{-2.1mm}
    \subfloat[]{
    \includegraphics[width=0.15\textwidth, height=0.05\textwidth]{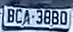}
    }\hspace{-2.1mm}
    \subfloat[]{
    \includegraphics[width=0.15\textwidth, height=0.05\textwidth]{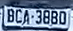}
    }
    }

    \vspace{-2.3mm}

    \resizebox{0.997\linewidth}{!}{
    \subfloat[]{
    \includegraphics[width=0.15\textwidth, height=0.05\textwidth]{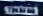}
    }\hspace{-2.1mm}
    \subfloat[]{
    \includegraphics[width=0.15\textwidth, height=0.05\textwidth]{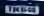}
    }\hspace{-2.1mm}
    \subfloat[]{
    \includegraphics[width=0.15\textwidth, height=0.05\textwidth]{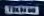}
    }\hspace{-2.1mm}
    \subfloat[]{
    \includegraphics[width=0.15\textwidth, height=0.05\textwidth]{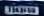}
    }\hspace{-2.1mm}
    \subfloat[]{
    \includegraphics[width=0.15\textwidth, height=0.05\textwidth]{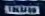}
    }\hspace{0.75mm}
    \subfloat[]{
    \includegraphics[width=0.15\textwidth, height=0.05\textwidth]{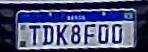}
    }\hspace{-2.1mm}
    \subfloat[]{
    \includegraphics[width=0.15\textwidth, height=0.05\textwidth]{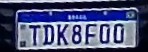}
    }\hspace{-2.1mm}
    \subfloat[]{
    \includegraphics[width=0.15\textwidth, height=0.05\textwidth]{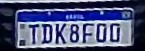}
    }\hspace{-2.1mm}
    \subfloat[]{
    \includegraphics[width=0.15\textwidth, height=0.05\textwidth]{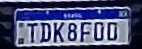}
    }\hspace{-2.1mm}
    \subfloat[]{
    \includegraphics[width=0.15\textwidth, height=0.05\textwidth]{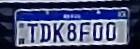}
    }
    }

    \vspace{-2.3mm}

    \resizebox{0.997\linewidth}{!}{
    \subfloat[]{
    \includegraphics[width=0.15\textwidth, height=0.05\textwidth]{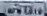}
    }\hspace{-2.1mm}
    \subfloat[]{
    \includegraphics[width=0.15\textwidth, height=0.05\textwidth]{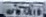}
    }\hspace{-2.1mm}
    \subfloat[]{
    \includegraphics[width=0.15\textwidth, height=0.05\textwidth]{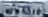}
    }\hspace{-2.1mm}
    \subfloat[]{
    \includegraphics[width=0.15\textwidth, height=0.05\textwidth]{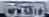}
    }\hspace{-2.1mm}
    \subfloat[]{
    \includegraphics[width=0.15\textwidth, height=0.05\textwidth]{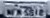}
    }\hspace{0.75mm}
    \subfloat[]{
    \includegraphics[width=0.15\textwidth, height=0.05\textwidth]{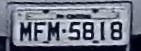}
    }\hspace{-2.1mm}
    \subfloat[]{
    \includegraphics[width=0.15\textwidth, height=0.05\textwidth]{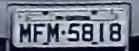}
    }\hspace{-2.1mm}
    \subfloat[]{
    \includegraphics[width=0.15\textwidth, height=0.05\textwidth]{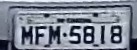}
    }\hspace{-2.1mm}
    \subfloat[]{
    \includegraphics[width=0.15\textwidth, height=0.05\textwidth]{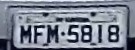}
    }\hspace{-2.1mm}
    \subfloat[]{
    \includegraphics[width=0.15\textwidth, height=0.05\textwidth]{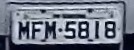}
    }
    }

    \vspace{-2.3mm}

    \resizebox{0.997\linewidth}{!}{
    \subfloat[]{
    \includegraphics[width=0.15\textwidth, height=0.05\textwidth]{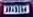}
    }\hspace{-2.1mm}
    \subfloat[]{
    \includegraphics[width=0.15\textwidth, height=0.05\textwidth]{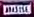}
    }\hspace{-2.1mm}
    \subfloat[]{
    \includegraphics[width=0.15\textwidth, height=0.05\textwidth]{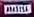}
    }\hspace{-2.1mm}
    \subfloat[]{
    \includegraphics[width=0.15\textwidth, height=0.05\textwidth]{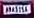}
    }\hspace{-2.1mm}
    \subfloat[]{
    \includegraphics[width=0.15\textwidth, height=0.05\textwidth]{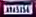}
    }\hspace{0.75mm}
    \subfloat[]{
    \includegraphics[width=0.15\textwidth, height=0.05\textwidth]{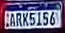}
    }\hspace{-2.1mm}
    \subfloat[]{
    \includegraphics[width=0.15\textwidth, height=0.05\textwidth]{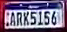}
    }\hspace{-2.1mm}
    \subfloat[]{
    \includegraphics[width=0.15\textwidth, height=0.05\textwidth]{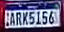}
    }\hspace{-2.1mm}
    \subfloat[]{
    \includegraphics[width=0.15\textwidth, height=0.05\textwidth]{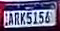}
    }\hspace{-2.1mm}
    \subfloat[]{
    \includegraphics[width=0.15\textwidth, height=0.05\textwidth]{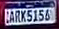}
    }
    }

    \vspace{-2.5mm}

    \caption{Examples of tracks from the training data. Each row corresponds to a complete track, showing five consecutive \acrfull*{lr} images on the left and five consecutive \acrfull*{hr} captures of the same \gls*{lp} on the right.}

    \label{fig:tracks-extracted}
\end{figure}

%% file: 2-figure-samples-original.tex
\begin{figure}[!htb]

    \centering

    \resizebox{0.975\linewidth}{!}{
        \includegraphics[width=0.23\linewidth]{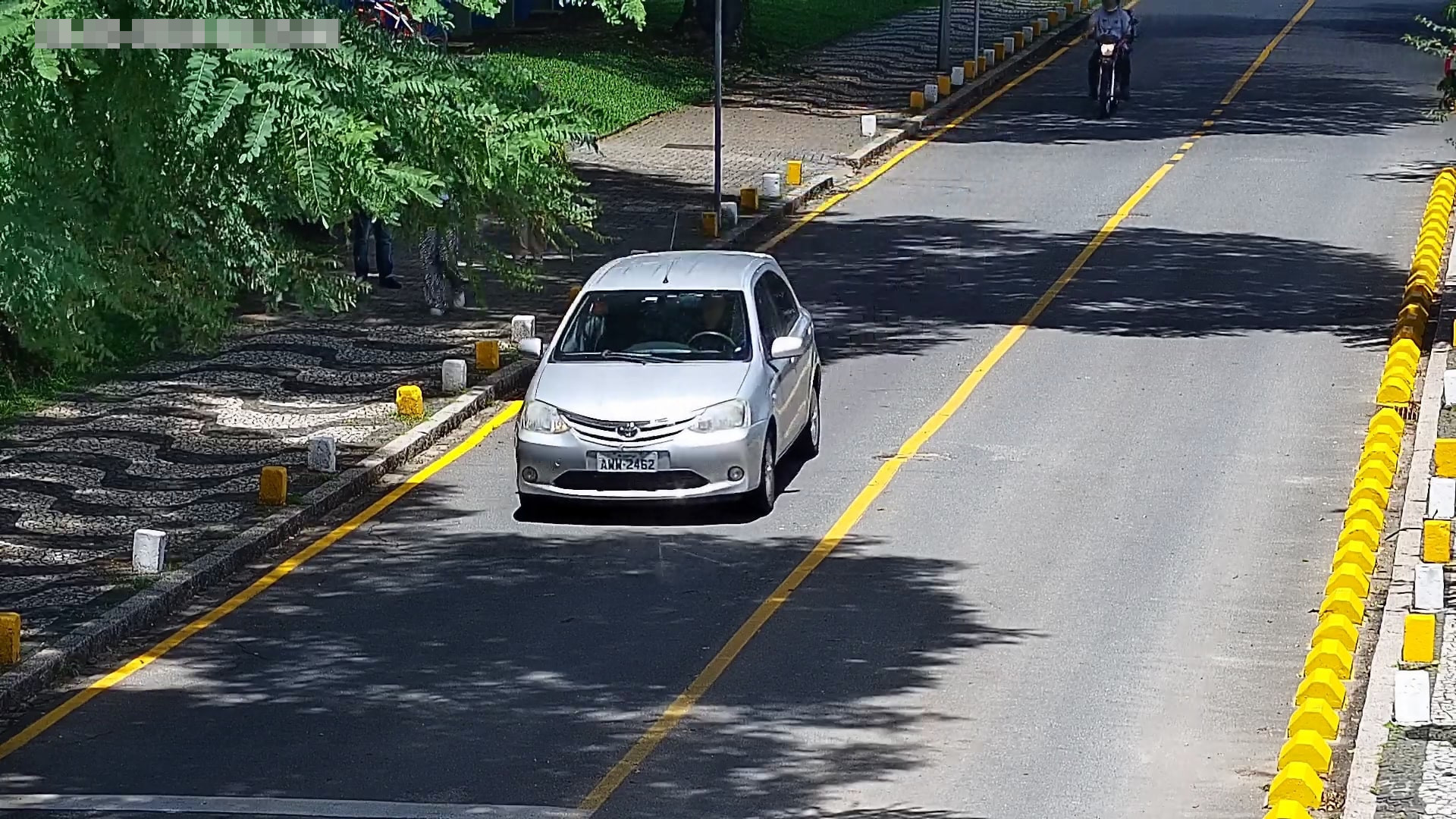} 
        \includegraphics[width=0.23\textwidth]{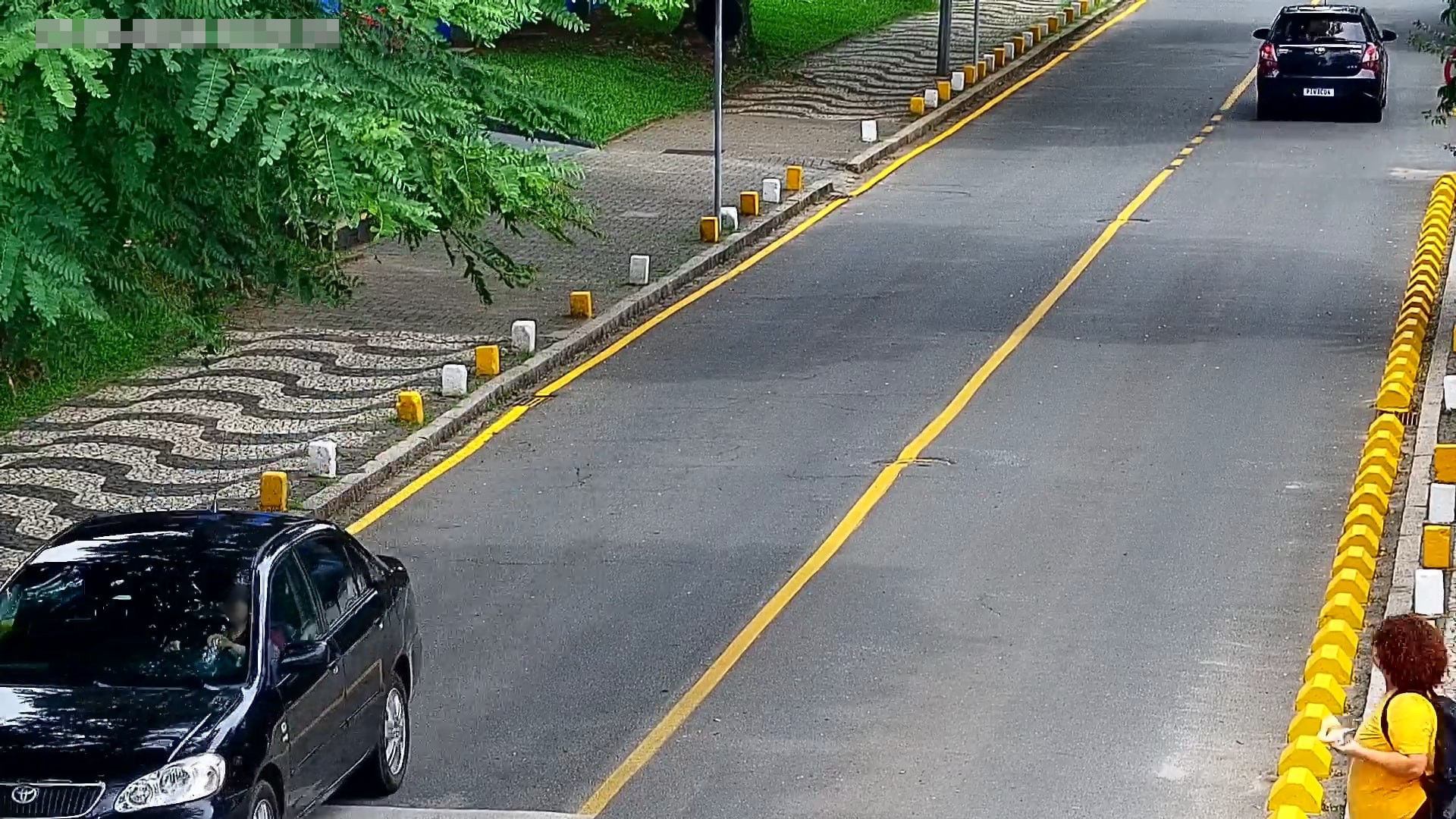}
        \includegraphics[width=0.23\textwidth]{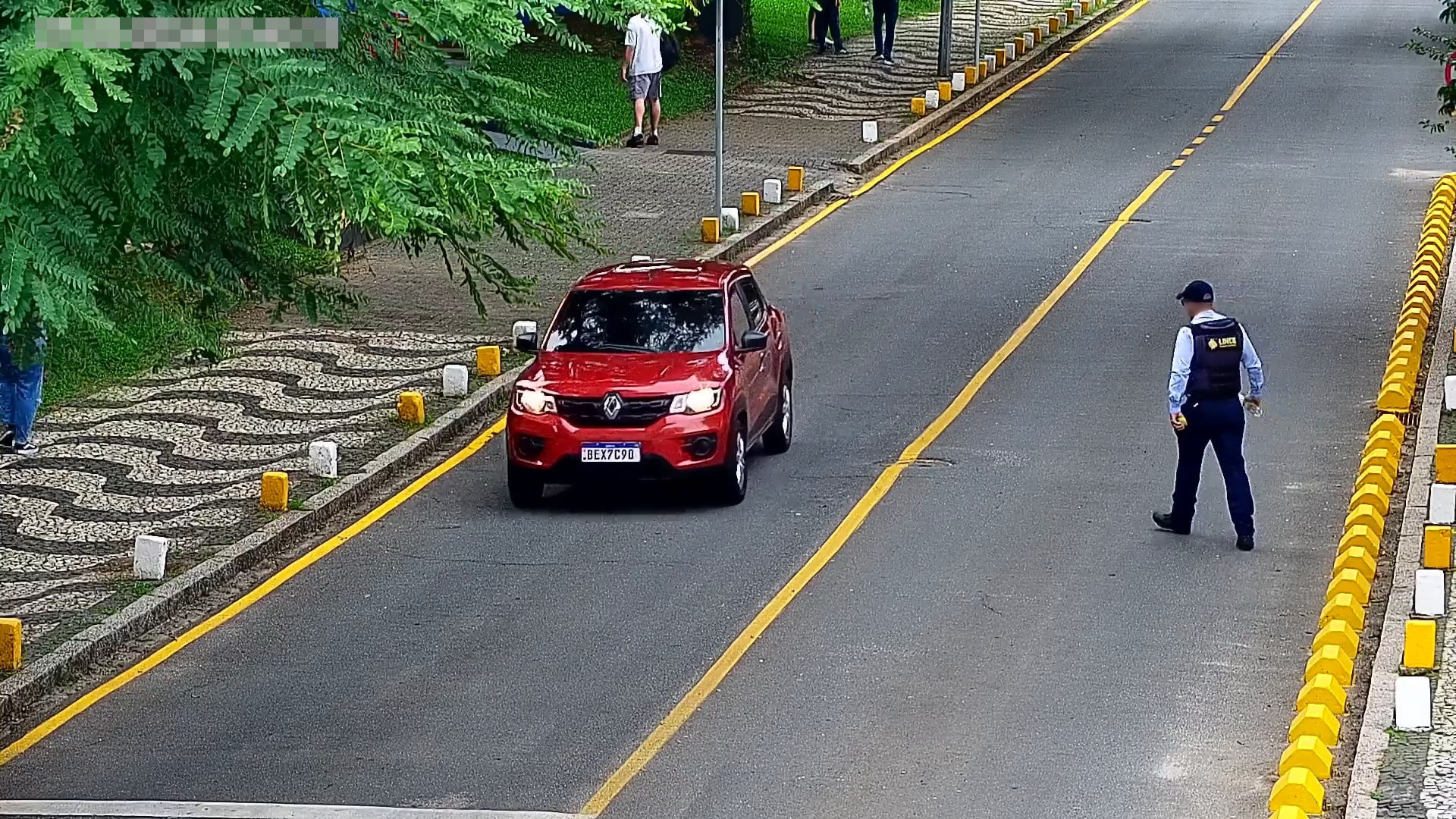}
        \includegraphics[width=0.23\textwidth]{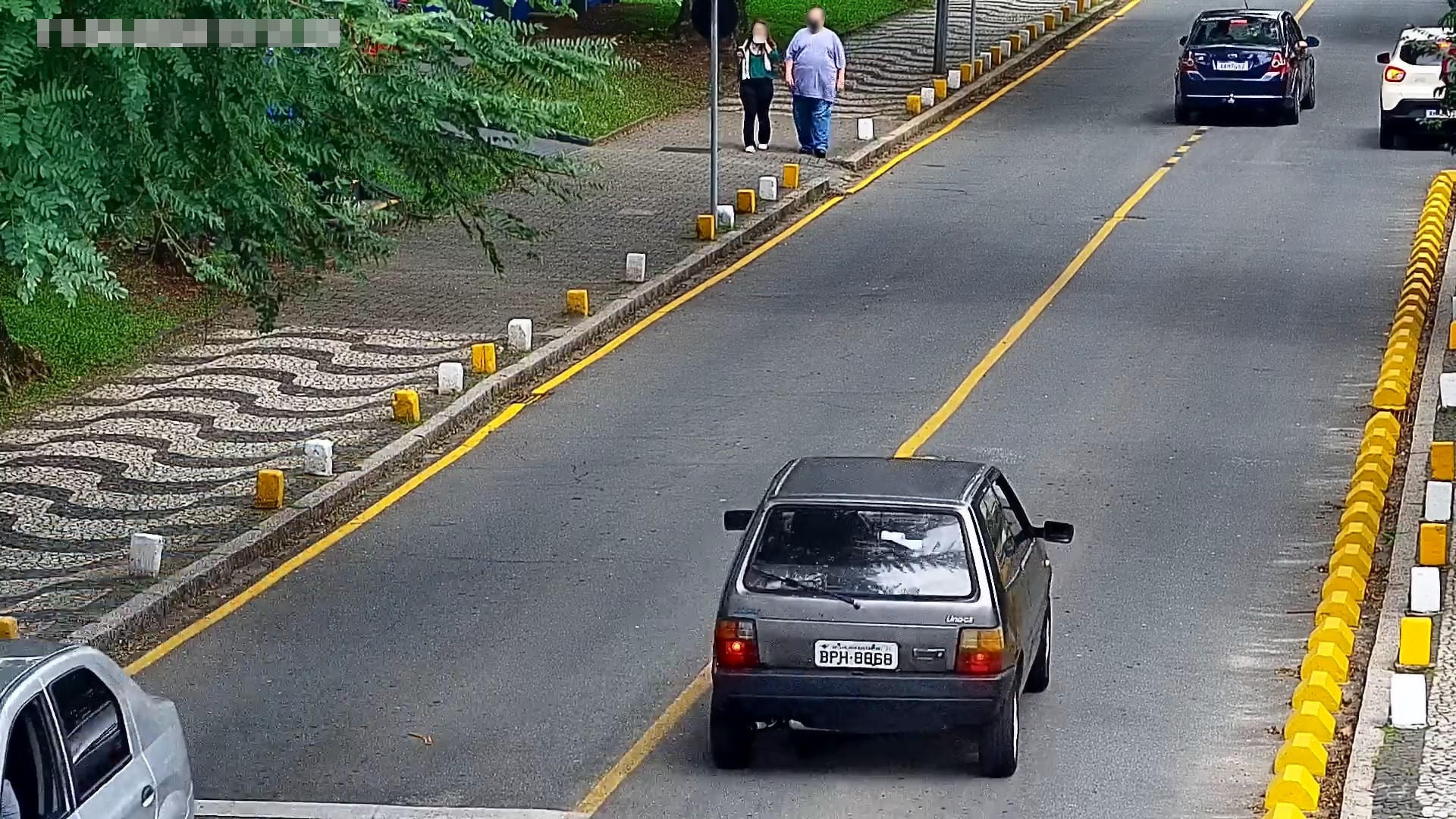}
    }
    
    \vspace{0.5mm}

    \resizebox{0.975\linewidth}{!}{
        \includegraphics[width=0.23\textwidth]{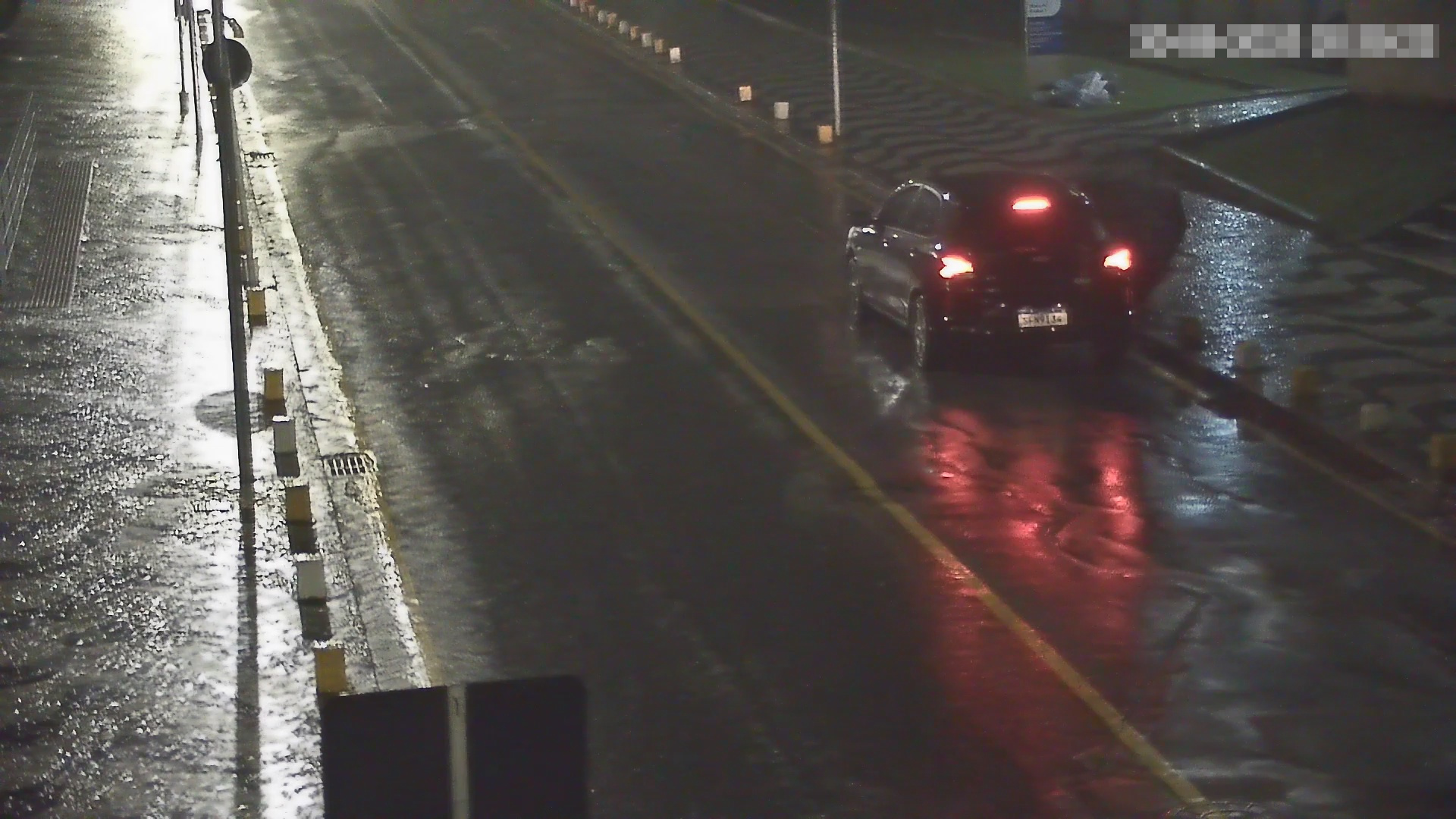}
        \includegraphics[width=0.23\textwidth]{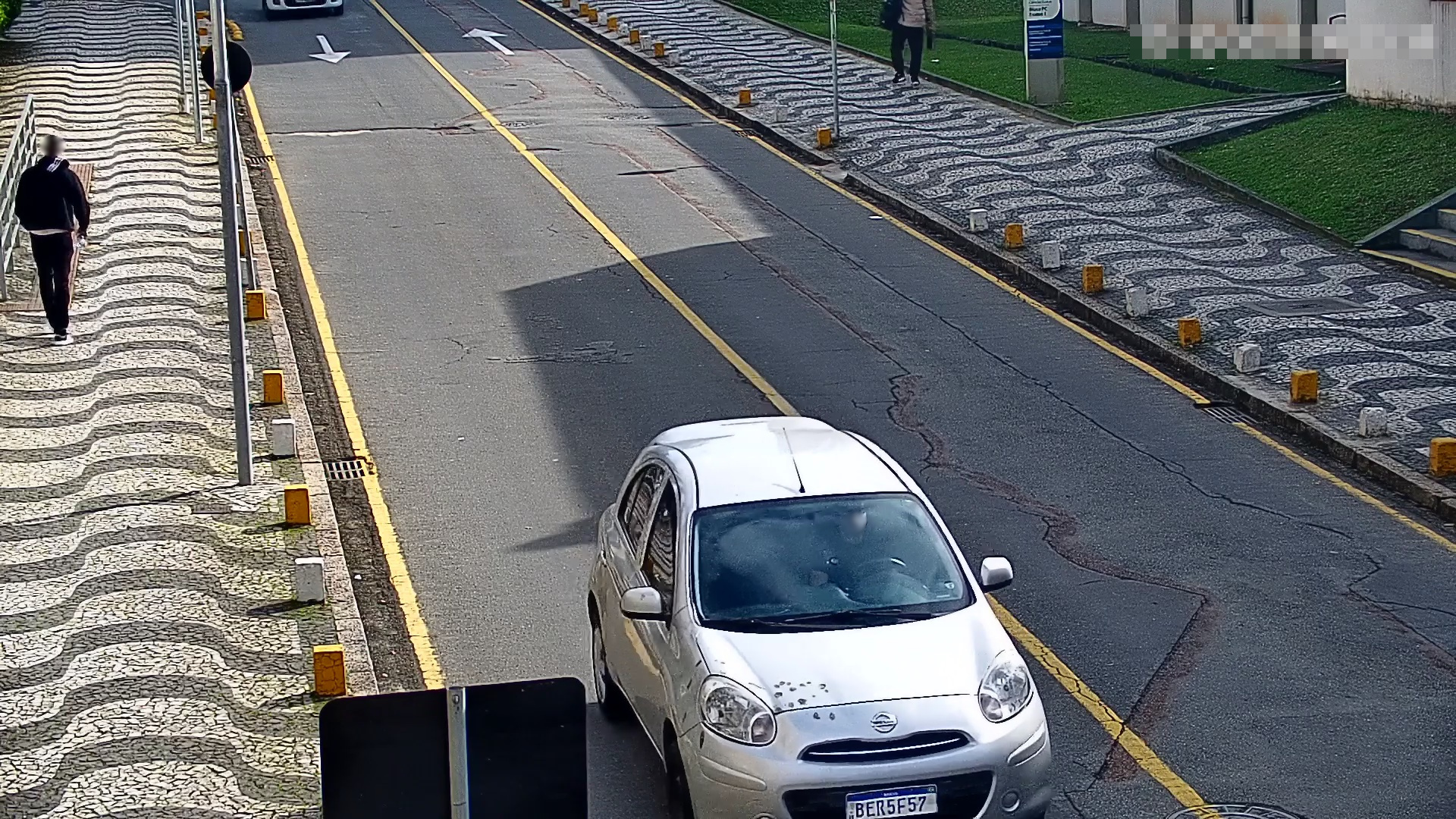}
        \includegraphics[width=0.23\textwidth]{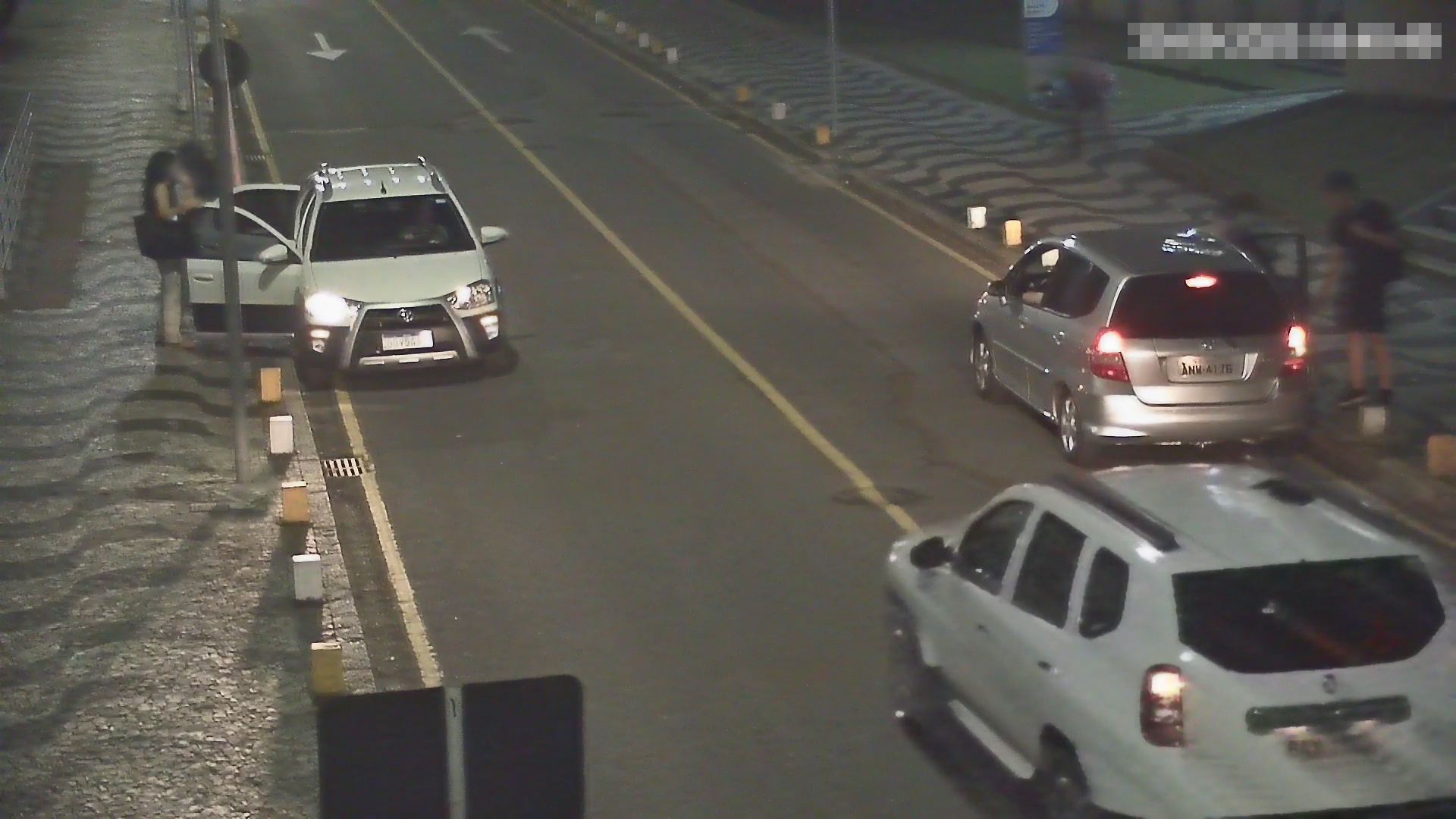}
        \includegraphics[width=0.23\textwidth]{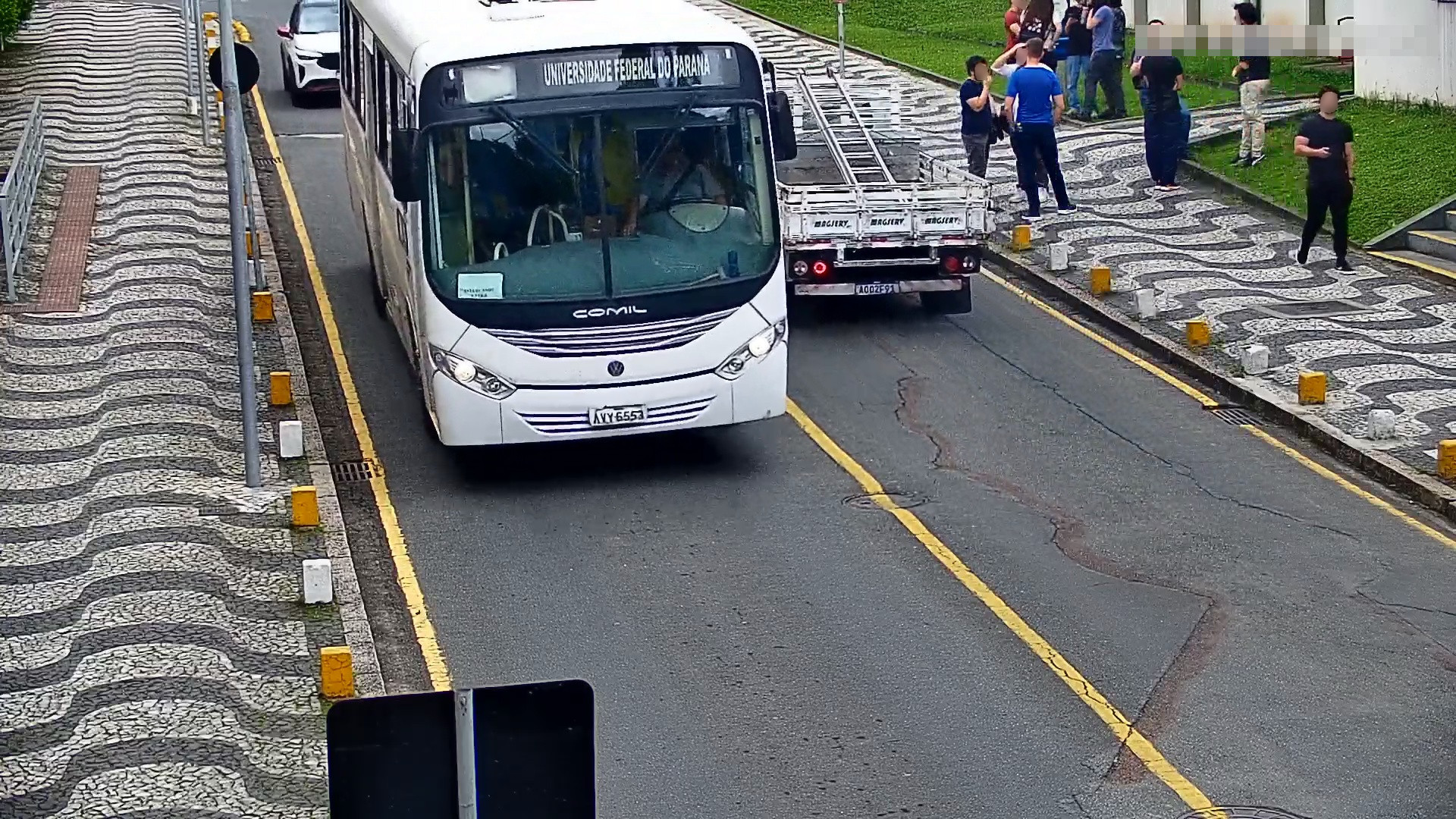}
    }

    \vspace{-2mm}
    
    \caption[]{Representative full-frame images (i.e., before \gls*{lp} detection and cropping) used to construct the \dataset dataset. The top row corresponds to Scenario~A, whereas the bottom row corresponds to Scenario~B, illustrating variations in vehicle categories and environmental conditions, including daylight, rain, and~nighttime.}

    \vspace{0.6mm}
        
    \label{fig:dataset-samples}
\end{figure}

%% file: 2-figure-pipeline.tex
\begin{figure}[!htb]
    \vspace{-0.5mm} %
    \centering
    \includegraphics[width=\linewidth]{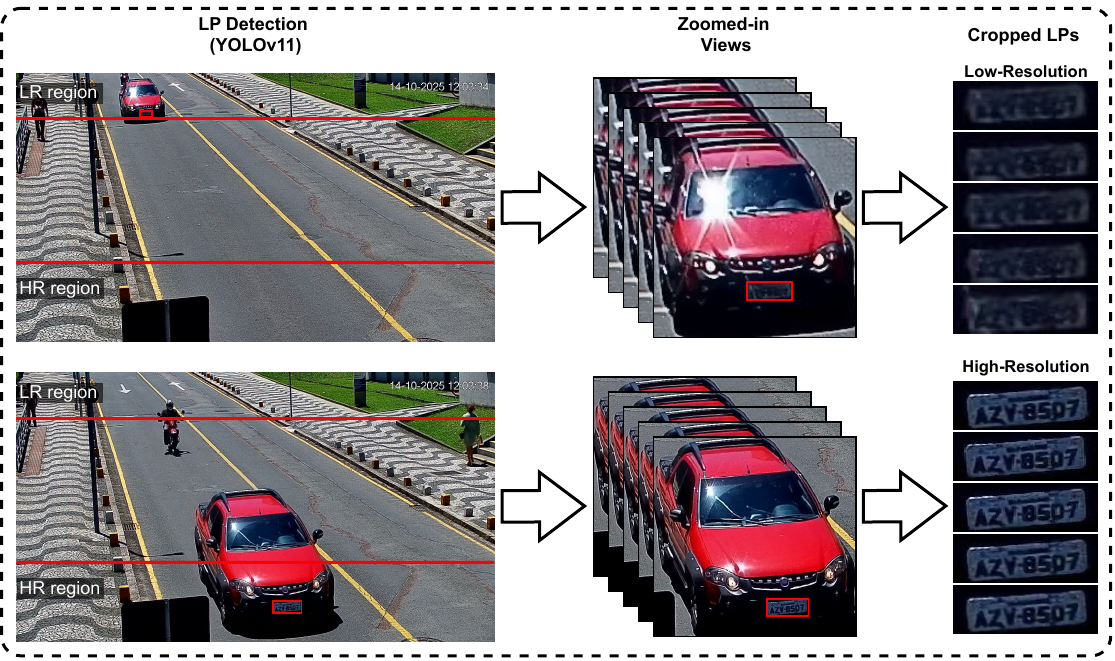}

    \vspace{-2mm}
    
    \caption[]{Overview of the dataset construction pipeline.
    \Glspl*{lp} are first detected using YOLOv11~\cite{yolov11} and then tracked across frames with BoT-SORT~\cite{aharon2022botsort}.
    For each vehicle, patches extracted from frames farther from the camera are used as \acrfull*{lr} samples, whereas patches from frames closer to the camera are used as \acrfull*{hr} samples. The final \gls*{lp} transcription is then obtained semi-automatically by applying an \gls*{ocr} model~\cite{goncalves2018realtime} to the \gls*{hr}~patches.}
    \label{fig:pipeline}
\end{figure}

%% file: 2-table-training-data.tex
\begin{table}[!htb]
\centering
\setlength{\tabcolsep}{5pt}
\caption{Distribution of training tracks by \gls*{lp} layout.}
\label{tab:training_data}

\vspace{-2mm}

\begin{tabular}{@{}lcccc@{}}
\toprule
 & \multicolumn{3}{c}{Number of Tracks} & \multirow{2}{*}{Unique LPs} \\ \cmidrule(lr){2-4}
 & Brazilian & Mercosur & Total &  \\ \midrule
Scenario A~\cite{nascimento2025toward} & 5,000 & 5,000 & 10,000 & \phantom{1}6,946 \\[0.25ex]
Scenario B~(new) & 2,000 & 8,000 & 10,000 & 10,000 \\ \bottomrule
\end{tabular}%
\end{table}

%% file: 3-results.tex
\section{Results}
\label{sec:results}

\begin{figure}[!htb]
    \centering

    \vspace{-1mm}
    
    \includegraphics[width=0.95\linewidth]{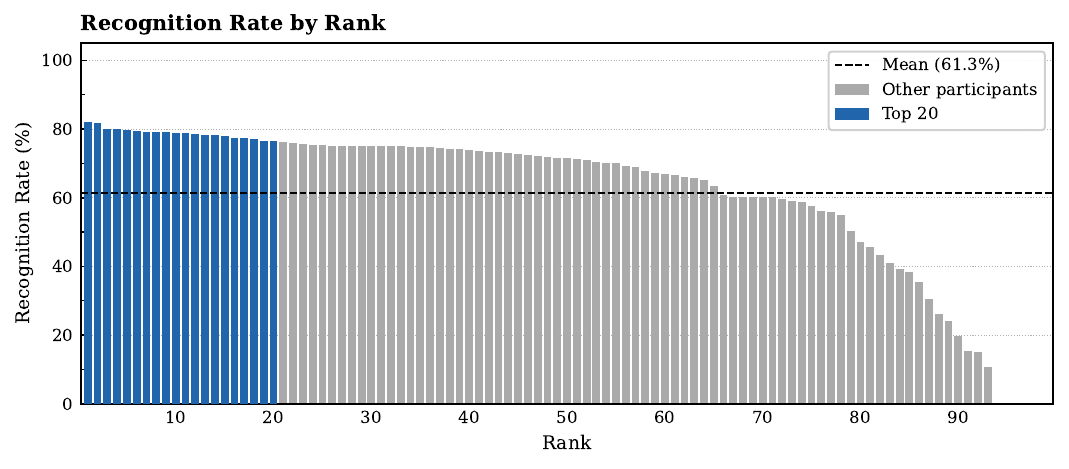}

    \vspace{-4mm}
    
    \caption{Recognition Rate achieved by all teams with valid submissions in the Blind Test Phase, sorted by final rank. The dashed horizontal line indicates the mean Recognition Rate across all participants, while the top 20 teams are highlighted in blue. Six teams~(ranked 94\textsuperscript{th} to 99\textsuperscript{th}) achieved Recognition Rates at or near 0\% and are therefore not visually distinguishable at this scale.}
    \label{fig:recognition-rate-vs-rank}
\end{figure}

\begin{figure}[!htb]
    \centering

    \includegraphics[width=0.675\linewidth]{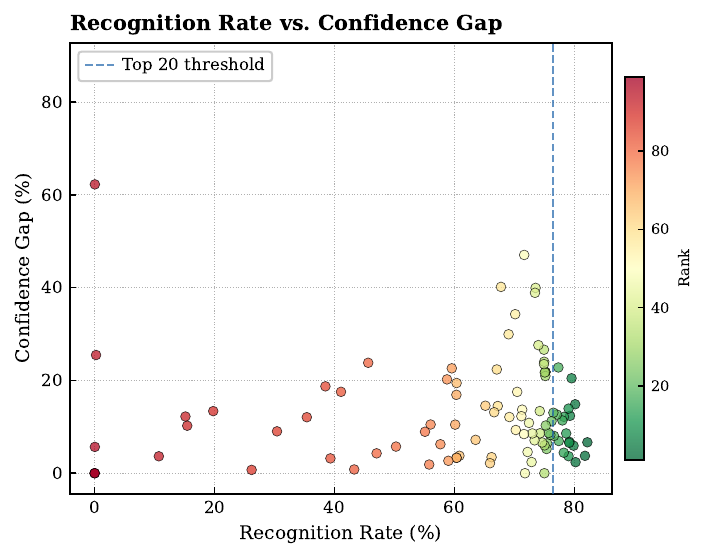}

    \vspace{-3.5mm}
    
    \caption{Recognition Rate versus Confidence Gap for all teams with valid submissions in the Blind Test Phase. The dashed vertical line indicates the Recognition Rate required to enter the top 20, and colors encode the final rank of each team.}
    \label{fig:recognition-rate-vs-confidence-gap-scatter}
\end{figure}

The results of the top 20 teams are presented in \cref{tab:results}, while \cref{fig:recognition-rate-vs-rank,fig:recognition-rate-vs-confidence-gap-scatter} provide a broader view of the performance distribution across all participating teams.
The complete leaderboard is available on the competition webpage.

\begin{table}[!htb]
    \centering
    \setlength{\tabcolsep}{6pt}
    \caption{Top 20 teams in the competition, ranked according to the official criterion, namely the Recognition Rate computed on the Blind Test Set. Confidence Gap is also reported, as it was used as a secondary criterion to break ties when needed.}
    \label{tab:results}

    \vspace{-2mm}

    \begin{tabular}{@{}c@{\hspace{1cm}}c@{}}
    \begin{tabular}{@{}c c c@{}}
    \toprule
    Rank & \begin{tabular}[c]{@{}c@{}}Recognition\\Rate~($\uparrow$)\end{tabular} & \begin{tabular}[c]{@{}c@{}}Confidence\\Gap~($\uparrow$)\end{tabular} \\
    \midrule
    1 & 82.13\% & \phantom{0}6.67\% \\
    2 & 81.73\% & \phantom{0}3.75\% \\
    3 & 80.17\% & \phantom{0}2.38\% \\
    4 & 80.10\% & 14.86\% \\
    5 & 79.83\% & \phantom{0}5.93\% \\
    6 & 79.50\% & 20.47\% \\
    7 & 79.23\% & 12.37\% \\
    8 & 79.13\% & \phantom{0}6.77\% \\
    9 & 79.10\% & \phantom{0}6.55\% \\
    10 & 79.00\% & 13.92\% \\
    \bottomrule
    \end{tabular}
    &
    \begin{tabular}{@{}c c c@{}}
    \toprule
    Rank & \begin{tabular}[c]{@{}c@{}}Recognition\\Rate~($\uparrow$)\end{tabular} & \begin{tabular}[c]{@{}c@{}}Confidence\\Gap~($\uparrow$)\end{tabular} \\
    \midrule
    11 & 79.00\% & \phantom{0}3.67\% \\
    12 & 78.60\% & \phantom{0}8.58\% \\
    13 & 78.23\% & 12.15\% \\
    14 & 78.20\% & \phantom{0}4.42\% \\
    15 & 77.97\% & 11.38\% \\
    16 & 77.37\% & \phantom{0}6.94\% \\
    17 & 77.30\% & 22.80\% \\
    18 & 77.07\% & 12.63\% \\
    19 & 76.57\% & \phantom{0}8.04\% \\
    20 & 76.47\% & 13.04\% \\
    \bottomrule
    \end{tabular}
    \end{tabular}
\end{table}

\cref{tab:results} shows that the competition was highly competitive at the top of the leaderboard.
The winning team achieved a Recognition Rate of 82.13\%, followed by 81.73\% and 80.17\% for the 2\textsuperscript{nd}- and 3\textsuperscript{rd}-ranked teams, respectively.
Notably, four teams surpassed the 80\% mark, and the top~10 teams were separated by only 3.13 percentage points.
Even when considering the entire top~20, the gap between the 1\textsuperscript{st}- and 20\textsuperscript{th}-ranked teams was only 5.66 percentage points, from 82.13\% to 76.47\%.
This narrow spread indicates that relatively small performance gains were sufficient to produce substantial changes in the final~ranking.

This strong competition at the top becomes even clearer in \cref{fig:recognition-rate-vs-rank}, which places the leading teams in the context of the full leaderboard.
All top~20 teams achieved a Recognition Rate of more than 76\%, whereas the overall mean across participants was 61.3\%.
The leftmost portion of the curve is notably flat, indicating that the highest-ranked teams were separated by relatively small margins despite achieving very strong performance.
More specifically, the recognition rate declines only slowly from the top positions through much of the ranking, and a substantial number of teams remained above the overall mean.
This pattern suggests that many teams were able to develop reasonably effective solutions, even if only a smaller group reached the highest performance band.
At the same time, \cref{fig:recognition-rate-vs-rank} underscores that \gls*{lrlpr} is far from a solved problem. Under the strict exact-match evaluation protocol standard in \gls*{alpr} research~\cite{silva2022flexible,laroca2023do,wei2024efficient}, even the top-performing method failed on 17.87\% of test~tracks.

\cref{fig:recognition-rate-vs-confidence-gap-scatter} provides a complementary perspective by jointly analyzing Recognition Rate and Confidence Gap.
The scatter plot shows that methods with similar recognition performance can differ substantially in Confidence Gap.
For instance, the 3\textsuperscript{rd}-ranked team achieved 80.17\% Recognition Rate with a Confidence Gap of only 2.38\%, whereas the 4\textsuperscript{th}- and 6\textsuperscript{th}-ranked teams obtained slightly lower Recognition Rates, namely 80.10\% and 79.50\%, but much higher Confidence Gaps of 14.86\% and 20.47\%, respectively.
A particularly illustrative example is the 17\textsuperscript{th}-ranked team, which achieved the highest Confidence Gap among the top~20, namely 22.80\%, despite a Recognition Rate of 77.30\%.

The following section discusses the top-ranked approaches in greater detail.

\subsection{Top-Ranked Approaches}
\label{sec:results:top_teams}

In this section, we briefly summarize the approaches of the top-5 teams in the competition.
The proposed methods differ substantially in terms of architecture, use of multiple frames, validation protocol, and reliance on external public~datasets.

\subsubsection{1\textsuperscript{st} Place}

\begin{figure}[!htb]
    \centering
    \includegraphics[width=0.9\linewidth]{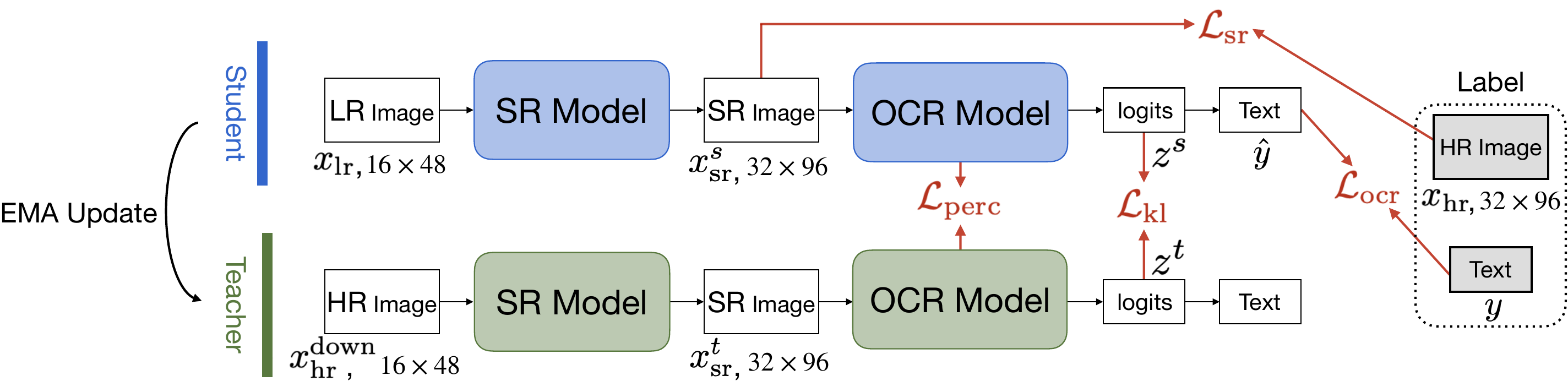}

    \vspace{-1mm}
    
    \caption{Overview of the teacher-student framework proposed by the winning team~(\textit{DLmath}). The student branch processes \acrfull*{lr} inputs while the teacher branch, updated via \acrfull*{ema}, receives downsampled \acrfull*{hr} images to guide the student’s learning.}
    \label{fig:top-01}
\end{figure}

The first-place team (\textit{DLmath}) from Korea University proposed a teacher-student framework that jointly trains a super-resolution model and an \gls*{ocr} model, as illustrated in \cref{fig:top-01}.
Specifically, the student branch takes \gls*{lr} inputs, whereas the teacher branch receives downsampled \gls*{hr} images; the teacher is initialized with weights pretrained on \gls*{hr} images only and is subsequently updated via \gls*{ema} of the student parameters.
Their loss combines reconstruction, recognition, perceptual, and KL-divergence~terms.

HATFIR~\cite{chen2026hat,zhang2023swinfir} and MambaIRv2~\cite{guo2025mambairv2} were adopted as super-resolution backbones, while GP-LPR~\cite{liu2024irregular} and a custom Transformer-based model were used for \gls*{ocr}.
During inference, the logits from all five \gls*{lr} frames within each track are summed before decoding.
The team further ensembled three independently trained models in the final submission by averaging their~logits.

To improve robustness to partial occlusions, the team applied vertical and horizontal masking augmentations during training.
From the official training data, 19,000 tracks were used for training, comprising 10,000 from Scenario-A and 9,000 from Scenario-B, while 1,000 additional Scenario-B samples were reserved for validation.
The training set was further expanded with three public datasets: OpenALPR-BR~\cite{openalpr_br}, RodoSol-ALPR~\cite{laroca2022cross}, and UFPR-ALPR~\cite{laroca2018robust}.

\subsubsection{2\textsuperscript{nd} Place}

The second-place team (\textit{AIO\_JiangnamCoffee}), composed of members from three Vietnamese institutions, namely the University of Information Technology, Ho Chi Minh University of Technology, and Vietnam National University, followed the four-stage framework of Baek~et~al.~\cite{baek2019what}.
Their pipeline consists of a \gls*{stn}~\cite{jaderberg2015spatial} for alignment, an SE\nobreakdash-ResNet\nobreakdash-C~\cite{hu2018squeeze,he2019bag} backbone for feature extraction, a Transformer encoder~\cite{vaswani2017attention} for sequence modeling, and a \gls*{ctc}~\cite{graves2006connectionist} head for prediction.
A CNN-based attention module estimates the quality of each frame and fuses the five input frames into a weighted representation.

\begin{figure}[!htb]
    \centering
    \includegraphics[width=0.975\linewidth]{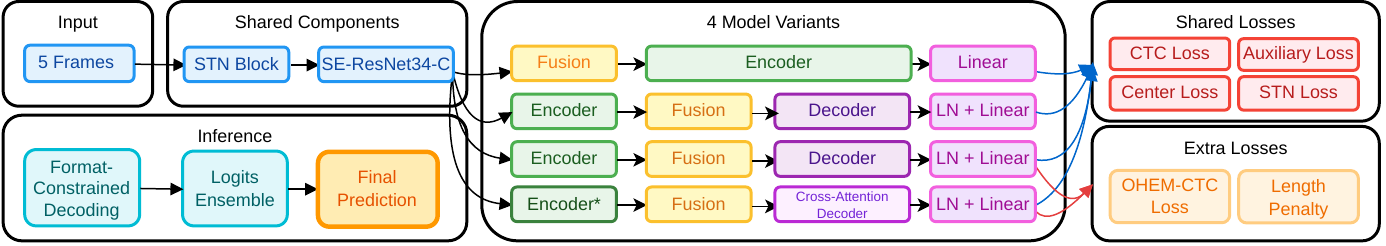}

    \vspace{-1.25mm}
    
    \caption{Overall ensemble architecture of the 2\textsuperscript{nd}-place team~(\textit{AIO\_JiangnamCoffee}). Five input frames are processed through a shared STN and SE-ResNet34 backbone and fed into four model variants that differ in Transformer encoder depth and the use of auxiliary losses (OHEM-CTC and length penalty).}
    \label{fig:top-02}
\end{figure}

Their final system ensembles four model variants that differ in the stage at which multi-frame fusion is performed and in their use of optional Transformer decoder components, as illustrated in \cref{fig:top-02}.
The training objective combines primary and auxiliary \gls*{ctc} losses, \gls*{stn} loss, center loss~\cite{olpadkar2021center}, and a length penalty, with \gls*{ohem}~\cite{shrivastava2016training} included in two of the models.
During inference, predictions are combined using weighted log-probability averaging, voting, and a single-model fallback strategy, while enforcing constraints associated with the Brazilian and Mercosur \gls*{lp}~layouts.

The models were trained from scratch for 30 epochs using AdamW~\cite{loshchilov2019decoupled_adamw} and a one-cycle learning rate schedule.
To effectively double the training set, \gls*{hr} frames were synthetically degraded using blur, noise, compression, and downscaling before standard augmentations such as affine and perspective transformations, color shifts, and coarse dropout were applied.
Only the official training data were used.
The validation set was drawn exclusively from Scenario-B, which is the scenario used in the test data, following a 90/10 train-validation~split.

\subsubsection{3\textsuperscript{rd} Place}

The third-place team (\textit{OpenOCR}), from the Institute of Trustworthy Embodied AI at Fudan University and the Shanghai Key Laboratory of Multimodal Embodied AI, both in China, treated \gls*{lrlpr} as a robust scene text recognition problem.
Rather than applying a dedicated super-resolution stage, each \gls*{lr} frame was directly fed to an \gls*{ocr} model, and the five per-frame predictions within a track were aggregated using a character-level voting strategy that combines both prediction frequency and per-character confidence~scores.

After comparing several candidate \gls*{ocr} models, including those proposed in~\cite{shi2017endtoend,bautista2022parseq,du2026mdiff4str}, the team adopted SVTRv2\nobreakdash-AR~\cite{du2025svtrv2,ye2026wrong} as their backbone, an attention-based autoregressive model whose visual encoder combines CNN and Transformer components.
Four SVTRv2-AR models were trained using two dataset configurations (official training data only vs.\ official training data augmented with \rodosol~\cite{laroca2022cross} and \ufpralpr~\cite{laroca2018robust}) and two initialization strategies (with and without Union14M\nobreakdash-L~\cite{jiang2023revisiting} and TextSSR~\cite{ye2025textssr} pretraining).
During inference, each track yields 20 predictions ($5~\text{frames} \times 4~\text{models}$), which are fused using character-level majority voting, with ties resolved by confidence~score.

Training used AdamW~\cite{loshchilov2019decoupled_adamw} with a one-cycle learning rate schedule over 100 epochs, including 10 warm-up epochs.
Data augmentation followed the PARSeq~\cite{bautista2022parseq} protocol.
A small validation set of 25 tracks~(125 images) was held out from the official training data to monitor overfitting.
The model contains 22.42M parameters and was trained on 8 NVIDIA V100~GPUs.

\subsubsection{4\textsuperscript{th} Place}

\begin{figure}[!htb]
    \centering
    \includegraphics[width=0.85\linewidth,
        trim={20mm 22mm 21mm 17mm}, %
        clip]{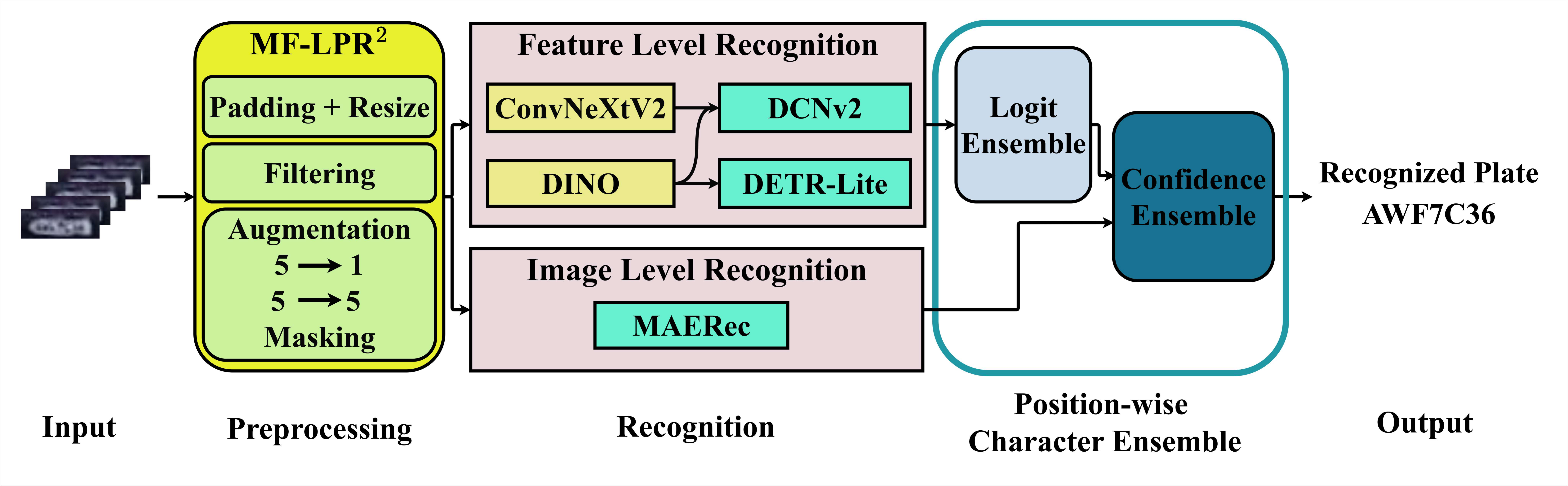}

    \vspace{-1.75mm}
    
    \caption{Overview of the pipeline proposed by the 4\textsuperscript{th}-place team~(\textit{CAP2}). It comprises MF-LPR$^2$-based preprocessing, dual-stream recognition with feature-level and image-level branches, and a two-stage position-wise character ensemble for final~prediction.}
    \label{fig:top-04}
\end{figure}

The fourth-place team (\textit{Capture And Predict Plate~--~CAP2}), from Handong Global University (Republic of Korea), proposed a multi-stage pipeline combining geometry-aware preprocessing, dual-stream recognition, and position-wise ensemble, as illustrated in \cref{fig:top-04}.
The preprocessing stage extends MF-LPR2~\cite{na2025mflpr2} with padding, resizing, filtering, and background suppression via U-Net~\cite{ronneberger2015unet}-generated text-region masks.
A key modification relative to the original MF-LPR2 is that, instead of fusing five frames into a single restored image~\mbox{($5\to1$)}, each frame is treated as an independent restored candidate~\mbox{($5\to5$)}.

Recognition combined multiple feature extractors (ConvNeXtV2~\cite{woo2023convnextv2}, DINOv2~\cite{oquab2024dinov2}, DINOv3~\cite{simeoni2025dinov3}) and recognizers (DETR-Lite~\cite{carion2020end}, DCNv2~\cite{zhu2019deformable}) at the feature level, alongside MAERec-B~\cite{jiang2023revisiting} at the image level.
Test-time augmentation averaged logits across brightness, contrast, and sharpness variants.
A two-stage position-wise character ensemble was then applied: first, a logit-level ensemble among the feature-level models using Optuna-optimized weights, followed by confidence-based fusion with the image-level~predictions.

All models were trained exclusively on the official training data, with 10\% held out for validation, using AdamW~\cite{loshchilov2019decoupled_adamw} with \textit{CosineWarmRestart} or \textit{CosineAnnealing} scheduling.
No external or synthetic data were used.

\subsubsection{5\textsuperscript{th} Place}

\begin{figure}[!htb]
    \centering
    \includegraphics[width=0.9\linewidth]{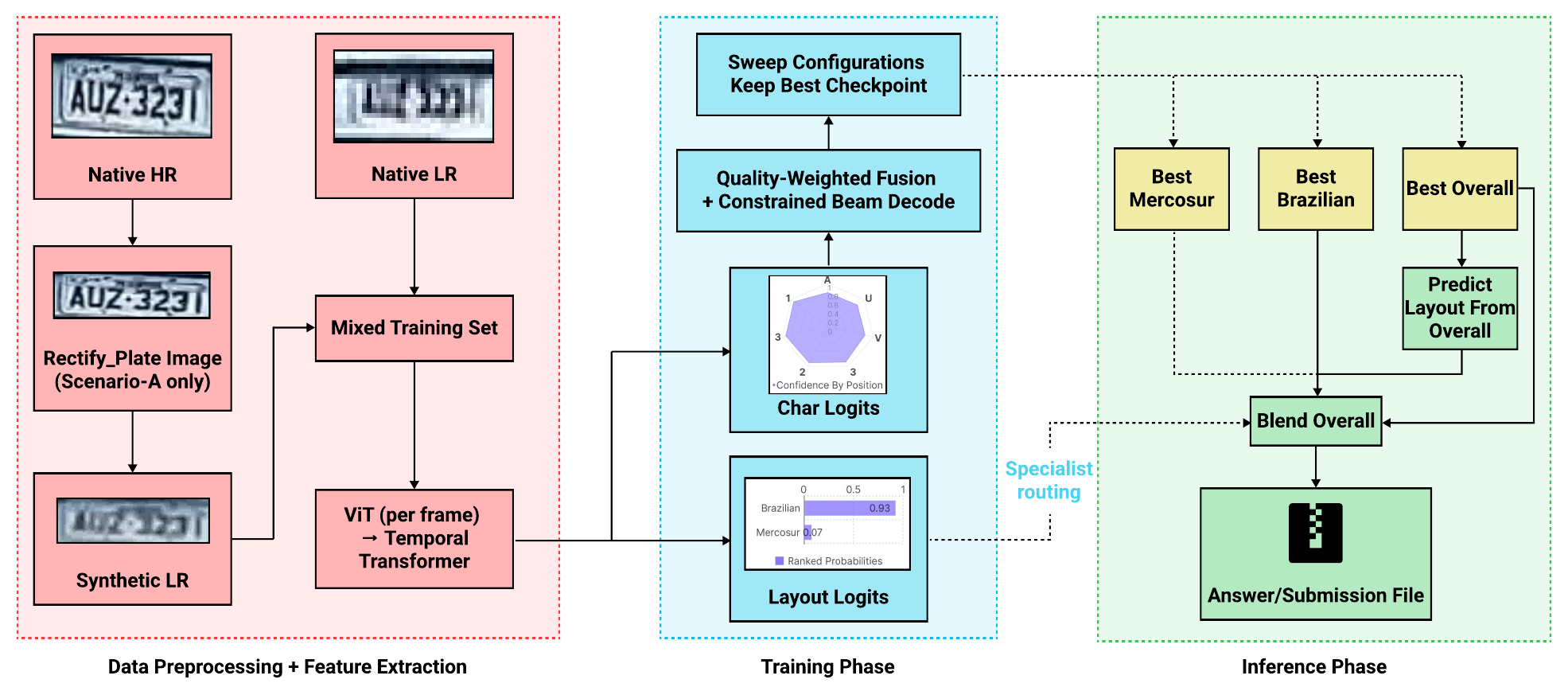}

    \vspace{-1.75mm}
    
    \caption{Overview of the pipeline proposed by the 5\textsuperscript{th}-place team~(\textit{UIT-MeoBeo}). Native \gls*{lr} images and synthetic \gls*{lr} samples generated from \gls*{hr} license plates were used to train a ViT-based multi-frame recognizer with dual heads for character and layout estimation. During inference, quality-weighted constrained decoding and layout-specific specialist routing were used to obtain the final~prediction.}
    \label{fig:top-05}
\end{figure}

The fifth-place team (\textit{UIT-MeoBeo}), from the University of Information Technology and Vietnam National University, both in Vietnam, proposed a multi-stage, multi-frame OCR pipeline combining geometry-aware preprocessing, Transformer-based recognition, and structure-aware decoding (see \cref{fig:top-05}).
The recognition backbone uses a PE-Core-L-14-336 frame encoder~\cite{hf_vit_pe_core_large_patch14_336} and a two-layer temporal Transformer for cross-frame fusion, with dual prediction heads jointly estimating the \gls*{lp} text and the layout (Brazilian or~Mercosur).

Key design choices include: (i)~structure-aware decoding with fixed 7-character output and position-wise letter/digit constraints; (ii)~layout-aware dual decoding, in which both Brazilian and Mercosur character masks are applied when the \gls*{lp} layout is uncertain, with the best hypothesis selected according to beam score and layout prior; (iii)~quality-aware multi-frame fusion using sharpness and noise proxies; and (iv)~a specialist cascade that routes predictions to layout-specific models when confidence is high.
Training mixed native \gls*{lr} images with synthetic \gls*{lr} samples generated from \gls*{hr} images via downscaling, upscaling, blur, JPEG compression, noise, and~sharpening.

Training used AdamW~\cite{loshchilov2019decoupled_adamw} with a cosine decay schedule and warmup~\cite{loshchilov2017sgdr}, incorporating gradient clipping, mixed precision~\cite{micikevicius2018mixed}, \gls*{ema} of model parameters, and early stopping.
Weighted sampling and layout-specific loss weighting were used to maintain class balance. The model contains approximately 300 million parameters and was trained on an NVIDIA RTX~4090~GPU.

\subsubsection{Approaches Ranked from 6\textsuperscript{th} to 10\textsuperscript{th}}

Among the teams ranked from 6\textsuperscript{th} to 10\textsuperscript{th}, only three submitted method summaries.
Although these notes are less detailed than those provided by the top-5 teams, they still offer useful insights into additional design choices explored in the~competition.

The 6\textsuperscript{th}-place team adopted a recognition-oriented strategy that avoided a full super-resolution stage, instead relying on lightweight enhancement, coarse alignment with an \gls*{stn}~\cite{jaderberg2015spatial}, HRNet-based~\cite{sun2019deep} multi-scale feature extraction, and a column-wise temporal fusion mechanism tailored to \gls*{ctc} decoding~\cite{graves2006connectionist}.
The 7\textsuperscript{th}-place team proposed an ensemble of heterogeneous recognizers, including ResNet+Transformer-based, Mamba-based~\cite{guo2025mambair,guo2025mambairv2}, and SVTR-based~\cite{du2022svtr,du2025svtrv2} models.
The 9\textsuperscript{th}-place team, in turn, introduced a pipeline combining spatial rectification, attention-based fusion of the five input frames, Transformer-based sequence modeling, and strategies such as \gls*{ema} and test-time~augmentation.

%% file: 4-discussion.tex
\section{Discussion}
\label{sec:discussion}

The results show that the competition was highly competitive and that the task remains clearly unsolved.
At the top of the leaderboard, the margins were very small: four teams exceeded 80\% Recognition Rate, the top 10 were separated by only 3.13 percentage points, and the gap between the 1\textsuperscript{st}- and 20\textsuperscript{th}-ranked teams was only 5.66 percentage points.
At the same time, the broader spread of scores and the winner's residual error rate of 17.87\% indicate that \gls*{lrlpr} remains challenging under the exact-match evaluation protocol.
Taken together, these findings suggest that the field has already developed strong solutions, but still has considerable room for improvement.

Another important takeaway is that there was no single dominant design among the best-performing methods.
Top teams adopted substantially different strategies, including explicit super-resolution, direct recognition from \gls*{lr} inputs, lightweight and heavy backbones, and different choices regarding the use of external public datasets.
What appears more consistent across the strongest submissions is the effective use of the five-frame track structure, whether through temporal fusion, voting, logit aggregation, or ensemble-based integration.
This indicates that future gains may depend less on committing to a specific architectural family and more on how well methods exploit complementary information across frames while respecting the structure of Brazilian and Mercosur~\glspl*{lp}.

The results also highlight \textit{Confidence Gap} as an important complementary target for future work.
As shown in \cref{fig:recognition-rate-vs-confidence-gap-scatter}, methods with similar Recognition Rates can differ substantially in Confidence Gap, which suggests that recognition accuracy and confidence quality capture different aspects of system behavior.
This distinction is relevant in practice, as confidence scores can support manual verification, candidate prioritization in forensic workflows, rejection of uncertain outputs, and the combination of evidence across frames.
Accordingly, improving \gls*{lrlpr} systems should involve not only increasing Recognition Rate, but also making confidence estimates more informative and better aligned with actual~correctness.

%% file: 5-conclusions.tex
\section{Conclusions}
\label{sec:conclusions}

In this paper, we presented the \textit{ICPR 2026 Competition on Low-Resolution License Plate Recognition}, covering the dataset, evaluation protocol, participation statistics, final results, and summaries of the top-ranked approaches.
To the best of our knowledge, this competition established the first large-scale international benchmark specifically focused on \gls*{lrlpr} with real low-quality data collected under operationally relevant conditions.
The participation of \numTeams teams from \numCountries countries also demonstrates the practical importance of the problem and the strong interest of the community in robust \gls*{alpr} under adverse conditions.

Beyond the final ranking itself, the competition provided a useful snapshot of the main methodological directions currently driving progress in the area.
The top submissions showed that strong performance can be achieved through different combinations of restoration, recognition, temporal modeling, ensembling, and layout-aware decoding.
This methodological diversity is one of the main outcomes of the benchmark, as it offers a clearer basis for future comparisons and for the design of stronger~baselines.

We expect this benchmark to support the next stage of research on robust \gls*{alpr}.
Promising directions include improved multi-frame modeling, better confidence estimation, stronger use of layout constraints, and tighter integration between restoration and recognition.
More broadly, the dataset, leaderboard, and method summaries may also stimulate follow-up studies on related problems such as \gls*{lp} super-resolution, \gls*{lp} legibility assessment, and forensic search support.
We hope this initiative helps advance recognition systems that are not only more accurate, but also more reliable in real-world applications.

%% file: 0-acknowledgments.tex
\section*{Acknowledgments}

This study was supported in part by the \textit{Coordenação de Aperfeiçoamento de Pessoal de Nível Superior} (CAPES) -- Finance Code 001, through the \textit{Programa de Excelência Acadêmica} (PROEX); in part by the \textit{Conselho Nacional de Desenvolvimento Científico e Tecnológico} (CNPq) under grant \#~315409/2023-1; and in part by the \textit{Fundação Araucária} under grant \#~078/2026. The authors also thank the \textit{Pontifícia Universidade Católica do Paraná}~(PUCPR) for the financial support that enabled conference participation.

%% file: bibtex.bib
@inproceedings{laroca2018robust,
  title = {A Robust Real-Time Automatic License Plate Recognition Based on the {YOLO} Detector},
  author = {R. {Laroca} and E. {Severo} and L. A. {Zanlorensi} and L. S. {Oliveira} and G. R. {Gon{\c{c}}alves} and W. R. {Schwartz} and D. {Menotti}},
  year = {2018},
  month = {July},
  booktitle = {International Joint Conference on Neural Networks (IJCNN)},
  volume = {},
  number = {},
  pages = {1-10},
  doi = {10.1109/IJCNN.2018.8489629},
  issn = {2161-4407}
}

@inproceedings{laroca2022cross,
  title = {On the Cross-Dataset Generalization in License Plate Recognition},
  author = {R. {Laroca} and E. V. {Cardoso} and D. R. {Lucio} and V. {Estevam} and D. {Menotti}},
  year = {2022},
  month = {Feb},
  booktitle = {International Conference on Computer Vision Theory and Applications (VISAPP)},
  volume = {},
  number = {},
  pages = {166-178},
  doi = {10.5220/0010846800003124},
  isbn = {978-989-758-555-5},
  issn = {2184-4321}
}

@inproceedings{laroca2023leveraging,
  title = {Leveraging Model Fusion for Improved License Plate Recognition},
  author = {R. {Laroca} and L. A. {Zanlorensi} and V. {Estevam} and R. {Minetto} and D. {Menotti}},
  year = {2023},
  month = {Nov},
  booktitle = {Iberoamerican Congress on Pattern Recognition (CIARP)},
  volume = {},
  number = {},
  pages = {60-75},
  doi = {10.1007/978-3-031-49249-5\_5},
  isbn = {978-3-031-49249-5}
}

@inproceedings{chen2025ntire_sr,
  title = {{NTIRE} 2025 Challenge on Image Super-Resolution ($\times 4$): Methods and Results},
  author = {Chen, Zheng and others},
  year = {2025},
  booktitle = {IEEE/CVF Conference on Computer Vision and Pattern Recognition Workshops (CVPRW)},
  volume = {},
  number = {},
  pages = {1516-1526},
  doi = {10.1109/CVPRW67362.2025.00141}
}

@inproceedings{coluccia2025drone,
  title = {The Drone-vs-Bird Detection Grand Challenge at {IJCNN} 2025},
  author = {Coluccia, Angelo and Fascista, Alessio and Dimou, Anastasios and Zarpalas, Dimitrios and Sommer, Lars and Schumann, Arne and Mele, Emanuele},
  year = {2025},
  booktitle = {International Joint Conference on Neural Networks (IJCNN)},
  volume = {},
  number = {},
  pages = {1-8},
  doi = {10.1109/IJCNN64981.2025.11228314}
}

@inproceedings{bautista2022parseq,
  title = {Scene Text Recognition with Permuted Autoregressive Sequence Models},
  author = {Bautista, Darwin and Atienza, Rowel},
  year = {2022},
  booktitle = {European Conference on Computer Vision (ECCV)},
  pages = {178-196},
  doi = {10.1007/978-3-031-19815-1_11},
  isbn = {978-3-031-19815-1}
}

@article{nascimento2024enhancing,
  title = {Enhancing License Plate Super-Resolution: A Layout-Aware and Character-Driven Approach},
  author = {V. {Nascimento} and R. {Laroca} and R. O. {Ribeiro} and W. R. {Schwartz} and D. {Menotti}},
  year = {2024},
  journal = {Conference on Graphics, Patterns and Images (SIBGRAPI)},
  volume = {},
  number = {},
  pages = {1-6},
  doi = {10.1109/SIBGRAPI62404.2024.10716303},
  issn = {1530-1834},
}

@article{oliveira2021vehicle,
  title = {{Vehicle-Rear}: A New Dataset to Explore Feature Fusion for Vehicle Identification Using Convolutional Neural Networks},
  author = {I. O. {Oliveira} and R. {Laroca} and D. {Menotti} and K. V. O. {Fonseca} and R. {Minetto}},
  year = {2021},
  journal = {IEEE Access},
  volume = {9},
  number = {},
  pages = {101065-101077},
  doi = {10.1109/ACCESS.2021.3097964},
  issn = {2169-3536}
}

@article{nascimento2025toward,
  title = {Toward Advancing License Plate Super-Resolution in Real-World Scenarios: A Dataset and Benchmark},
  author = {V. {Nascimento} and G. E. {Lima} and R. O. {Ribeiro} and W. R. {Schwartz} and R. {Laroca} and D. {Menotti}},
  year = {2025},
  journal = {Journal of the Brazilian Computer Society},
  volume = {1},
  number = {31},
  pages = {435-449},
  doi = {10.5753/jbcs.2025.5159},
  issn = {},
}

@misc{yolov11,
  title = {{YOLOv11}},
  author = {Ultralytics},
  year = {2026},
  url = {https://docs.ultralytics.com/models/yolo11/},
  note = {Accessed: 2026-03-30}
}

@article{aharon2022botsort,
  title={{BoT-SORT}: Robust Associations Multi-Pedestrian Tracking},
  author={Aharon, Nir and Orfaig, Roy and Bobrovsky, Ben-Zion},
  journal = {arXiv preprint arXiv:2104.07636},
  year={2022},
  number={},
  pages = {1-13}
}

@inproceedings{goncalves2018realtime,
  title = {Real-Time Automatic License Plate Recognition through Deep Multi-Task Networks},
  author = {G. R. {Gon{\c{c}}alves} and M. A. {Diniz} and R. {Laroca} and D. {Menotti} and W. R. {Schwartz}},
  year = {2018},
  month = {Oct},
  booktitle = {Conference on Graphics, Patterns and Images (SIBGRAPI)},
  volume = {},
  number = {},
  pages = {110-117},
  doi = {10.1109/SIBGRAPI.2018.00021},
  issn = {2377-5416}
}

@inproceedings{laroca2025improving,
  year = {2025},
  title = {Improving Small Drone Detection Through Multi-Scale Processing and Data Augmentation},
  author = {R. {Laroca} and M. {dos Santos} and D. {Menotti}},
  month = {June},
  booktitle = {International Joint Conference on Neural Networks (IJCNN)},
  pages = {1-8},
  doi={10.1109/IJCNN64981.2025.11227421}
}

@article{diwan2023object,
  title        = {Object detection using {YOLO}: challenges, architectural successors, datasets and applications},
  author       = {Diwan, Tausif and Anirudh, G. and Tembhurne, Jitendra V.},
  year         = {2023},
  month        = {},
  day          = {},
  journal      = {Multimedia Tools and Applications},
  volume       = {82},
  number       = {6},
  pages        = {9243-9275},
  doi          = {10.1007/s11042-022-13644-y},
  issn         = {1573-7721}
}

@article{ismail2025automatic,
  title = {Automatic License Plate Recognition in In-the-Wild Scenarios: A Comprehensive Review, Open Issues, and Future Directions},
  author = {Ismail, Amir and Mehri, Maroua and Sahbani, Anis and Essoukri Ben Amara, Najoua},
  year = {2025},
  journal = {IEEE Access},
  volume = {13},
  number = {},
  pages = {145387-145415},
  doi = {10.1109/ACCESS.2025.3598971}
}

@article{ke2023ultra,
  title = {An Ultra-Fast Automatic License Plate Recognition Approach for Unconstrained Scenarios},
  author = {Xiao {Ke} and Zeng, Ganxiong and Guo, Wenzhong},
  year = {2023},
  journal = {IEEE Transactions on Intelligent Transportation Systems},
  volume = {24},
  number = {5},
  pages = {5172-5185},
  doi = {10.1109/TITS.2023.3237581}
}

@article{laroca2025advancing,
  title = {Advancing Multinational License Plate Recognition Through Synthetic and Real Data Fusion: A Comprehensive Evaluation},
  author = {{Laroca}, Rayson and {Estevam}, Valter and {Moreira}, Gladston J. P. and {Minetto}, Rodrigo and {Menotti}, David},
  year = {2025},
  journal = {IET Intelligent Transport Systems},
  volume = {19},
  number = {1},
  pages = {e70086},
  doi = {10.1049/itr2.70086}
}

@article{rao2024license,
  title = {License plate recognition system in unconstrained scenes via a new image correction scheme and improved {CRNN}},
  author = {Zhan {Rao} and Dezhi Yang and Ning Chen and Jian Liu},
  year = {2024},
  journal = {Expert Systems with Applications},
  volume = {243},
  pages = {122878},
  doi = {10.1016/j.eswa.2023.122878},
  issn = {0957-4174}
}

@inproceedings{gong2024dataset,
  title = {A dataset and model for realistic license plate deblurring},
  author = {Gong, Haoyan and Feng, Yuzheng and Zhang, Zhenrong and Hou, Xianxu and Liu, Jingxin and Huang, Siqi and Liu, Hongbin},
  year = {2024},
  booktitle = {International Joint Conference on Artificial Intelligence (IJCAI)},
  pages = {1-9},
  doi = {10.24963/ijcai.2024/86},
  isbn = {978-1-956792-04-1}
}

@inproceedings{gong2025lpdiff,
  title = {{LP-Diff}: Towards Improved Restoration of Real-World Degraded License Plate},
  author = {Gong, Haoyan and Zhang, Zhenrong and Feng, Yuzheng and Nguyen, Anh and Liu, Hongbin},
  year = {2025},
  booktitle = {IEEE/CVF Conference on Computer Vision and Pattern Recognition (CVPR)},
  volume = {},
  number = {},
  pages = {17831-17840},
  doi = {10.1109/CVPR52734.2025.01661}
}

@article{silva2022flexible,
  title = {A Flexible Approach for Automatic License Plate Recognition in Unconstrained Scenarios},
  author = {S. M. {Silva} and C. R. {Jung}},
  year = {2022},
  journal = {IEEE Transactions on Intelligent Transportation Systems},
  volume = {23},
  number = {6},
  pages = {5693-5703},
  doi = {10.1109/TITS.2021.3055946}
}

@inproceedings{goncalves2019multitask,
  title = {Multi-Task Learning for Low-Resolution License Plate Recognition},
  author = {G. R. {Gon{\c{c}}alves} and M. A. {Diniz} and R. {Laroca} and D. {Menotti} and W. R. {Schwartz}},
  year = {2019},
  month = {Oct},
  booktitle = {Iberoamerican Congress on Pattern Recognition (CIARP)},
  volume = {},
  number = {},
  pages = {251-261},
  doi = {10.1007/978-3-030-33904-3\_23},
  isbn = {978-3-030-33904-3}
}

@article{pan2024lpsrgan,
  title = {{LPSRGAN}: Generative adversarial networks for super-resolution of license plate image},
  author = {Yuecheng Pan and Jin Tang and Tardi Tjahjadi},
  year = {2024},
  journal = {Neurocomputing},
  volume = {580},
  pages = {127426},
  doi = {10.1016/j.neucom.2024.127426},
  issn = {0925-2312}
}

@article{schirrmacher2023benchmarking,
  title = {Benchmarking Probabilistic Deep Learning Methods for License Plate Recognition},
  author = {Franziska {Schirrmacher} and Lorch, Benedikt and Maier, Anatol and Riess, Christian},
  year = {2023},
  journal = {IEEE Transactions on Intelligent Transportation Systems},
  volume = {24},
  number = {9},
  pages = {9203-9216},
  doi = {10.1109/TITS.2023.3278533}
}

@inproceedings{moussa2022forensic,
  title = {Forensic License Plate Recognition with Compression-Informed Transformers},
  author = {Denise {Moussa} and Maier, Anatol and Spruck, Andreas and Seiler, Jürgen and Riess, Christian},
  year = {2022},
  booktitle = {IEEE International Conference on Image Processing (ICIP)},
  volume = {},
  number = {},
  month = {Oct},
  pages = {406-410},
  doi = {10.1109/ICIP46576.2022.9897178}
}

@article{kim2024afanet,
  year = {2024},
  title = {{AFA-Net}: Adaptive Feature Attention Network in image deblurring and super-resolution for improving license plate recognition},
  author = {Dogun {Kim} and Jin Kim and Eunil Park},
  journal = {Computer Vision and Image Understanding},
  volume = {238},
  pages = {103879},
  doi = {10.1016/j.cviu.2023.103879},
  issn = {1077-3142}
}

@inproceedings{nascimento2022combining,
  title = {Combining Attention Module and Pixel Shuffle for License Plate Super-resolution},
  author = {V. {Nascimento} and R. {Laroca} and J. A. {Lambert} and W. R. {Schwartz} and D. {Menotti}},
  year = {2022},
  month = {Oct},
  booktitle = {Conference on Graphics, Patterns and Images (SIBGRAPI)},
  volume = {},
  number = {},
  pages = {228-233},
  doi = {10.1109/SIBGRAPI55357.2022.9991753},
  issn = {1530-1834},
}

@article{laroca2021efficient,
  title = {An Efficient and Layout-Independent Automatic License Plate Recognition System Based on the {YOLO} Detector},
  author = {R. {Laroca} and L. A. {Zanlorensi} and G. R. {Gon{\c{c}}alves} and E. {Todt} and W. R. {Schwartz} and D. {Menotti}},
  year = {2021},
  journal = {IET Intelligent Transport Systems},
  volume = {15},
  number = {4},
  pages = {483-503},
  doi = {10.1049/itr2.12030},
  issn = {1751-956X}
}

@inproceedings{maier2022reliability,
  title = {Reliability Scoring for the Recognition of Degraded License Plates},
  author = {Anatol {Maier} and Moussa, Denise and Spruck, Andreas and Seiler, J\"{u}rgen and Riess, Christian},
  year = {2022},
  booktitle = {IEEE International Conference on Advanced Video and Signal Based Surveillance (AVSS)},
  volume = {},
  number = {},
  pages = {1-8},
  doi = {10.1109/AVSS56176.2022.9959390}
}

@article{codabench,
title = {Codabench: Flexible, easy-to-use, and reproducible meta-benchmark platform},
author = {Zhen Xu and Sergio Escalera and Adrien Pavão and Magali Richard and
Wei-Wei Tu and Quanming Yao and Huan Zhao and Isabelle Guyon},
journal = {Patterns},
volume = {3},
number = {7},
pages = {100543},
year = {2022},
issn = {2666-3899},
doi = {10.1016/j.patter.2022.100543},
}

@article{chen2026hat,
  title = {{HAT}: Hybrid Attention Transformer for Image Restoration},
  author = {Chen, Xiangyu and Wang, Xintao and Zhang, Wenlong and Kong, Xiangtao and Qiao, Yu and Zhou, Jiantao and Dong, Chao},
  year = {2026},
  journal = {IEEE Transactions on Pattern Analysis and Machine Intelligence},
  volume = {48},
  number = {3},
  pages = {2676-2694},
  doi = {10.1109/TPAMI.2025.3628275}
}

@article{zhang2023swinfir,
  title = {{SwinFIR}: Revisiting the {SwinIR} with Fast Fourier Convolution and Improved Training for Image Super-Resolution},
  author = {Dafeng Zhang and Feiyu Huang and Shizhuo Liu and Xiaobing Wang and Zhezhu Jin},
  year = {2023},
  journal = {arXiv preprint arXiv:2208.11247},
  number = {},
  pages = {1-14}
}

@inproceedings{guo2025mambair,
  title = {{MambaIR}: A Simple Baseline for Image Restoration with State-Space Model},
  author = {Guo, Hang and Li, Jinmin and Dai, Tao and Ouyang, Zhihao and Ren, Xudong and Xia, Shu-Tao},
  year = {2025},
  booktitle = {European Conference on Computer Vision (ECCV)},
  pages = {222-241},
  doi = {10.1007/978-3-031-72649-1_13},
  isbn = {978-3-031-72649-1}
}

@inproceedings{guo2025mambairv2,
  title = {{MambaIRv2}: Attentive State Space Restoration},
  author = {Guo, Hang and Guo, Yong and Zha, Yaohua and Zhang, Yulun and Li, Wenbo and Dai, Tao and Xia, Shu-Tao and Li, Yawei},
  year = {2025},
  booktitle = {IEEE/CVF Conference on Computer Vision and Pattern Recognition (CVPR)},
  volume = {},
  number = {},
  pages = {28124-28133},
  doi = {10.1109/CVPR52734.2025.02619}
}

@misc{openalpr_br,
  title = {{OpenALPR-BR dataset}},
  author = {{OpenALPR}},
  year = {2016},
  howpublished = {\url{https://github.com/openalpr/benchmarks/tree/master/endtoend/br}}
}

@article{na2025mflpr2,
  title = {{MF-LPR2}: Multi-frame license plate image restoration and recognition using optical flow},
  author = {Kihyun Na and Junseok Oh and Youngkwan Cho and Bumjin Kim and Sungmin Cho and Jinyoung Choi and Injung Kim},
  year = {2025},
  journal = {Computer Vision and Image Understanding},
  volume = {256},
  pages = {104361},
  doi = {10.1016/j.cviu.2025.104361},
  issn = {1077-3142}
}

@inproceedings{baek2019what,
  title = {What Is Wrong With Scene Text Recognition Model Comparisons? Dataset and Model Analysis},
  author = {Jeonghun {Baek} and Kim, Geewook and Lee, Junyeop and Park, Sungrae and Han, Dongyoon and Yun, Sangdoo and Oh, Seong Joon and Lee, Hwalsuk},
  year = {2019},
  booktitle = {IEEE/CVF International Conference on Computer Vision (ICCV)},
  volume = {},
  number = {},
  pages = {4714-4722},
  doi = {10.1109/ICCV.2019.00481}
}

@InProceedings{du2025svtrv2,
    author    = {Du, Yongkun and Chen, Zhineng and Xie, Hongtao and Jia, Caiyan and Jiang, Yu-Gang},
    title     = {{SVTRv2}: {CTC} Beats Encoder-Decoder Models in Scene Text Recognition},
    booktitle = {IEEE/CVF International Conference on Computer Vision~(ICCV)},
    month     = {Oct},
    year      = {2025},
    pages     = {20147-20156}
}

@inproceedings{ye2026wrong,
  title={What's Wrong with Synthetic Data for Scene Text Recognition? A Strong Synthetic Engine with Diverse Simulations and Self-Evolution},
  author={Ye, Xingsong and Du, Yongkun and Zhang, JiaXin and Li, Chen and LYU, Jing and Chen, Zhineng},
  booktitle = {IEEE/CVF Conference on Computer Vision and Pattern Recognition (CVPR)},
  year={2026}
}

@misc{hf_vit_pe_core_large_patch14_336,
  author = {{Hugging Face}},
  title = {Model card and pretrained weights for PE-Core-L-14-336},
  howpublished = {Hugging Face model hub},
  note = {timm-remapped image-encoder-only variant},
  year = {2025}
}

@inproceedings{jiang2023revisiting,
  title = {Revisiting Scene Text Recognition: A Data Perspective},
  author = {Qing {Jiang} and Wang, Jiapeng and Peng, Dezhi and Liu, Chongyu and Jin, Lianwen},
  year = {2023},
  booktitle = {IEEE/CVF International Conference on Computer Vision~(ICCV)},
  volume = {},
  number = {},
  pages = {20486-20497},
  doi = {10.1109/ICCV51070.2023.01878}
}

@inproceedings{ye2025textssr,
  title = {{TextSSR}: Diffusion-based Data Synthesis for Scene Text Recognition},
  author = {Xingsong Ye and Yongkun Du and Yunbo Tao and Zhineng Chen},
  year = {2025},
  booktitle = {IEEE/CVF International Conference on Computer Vision (ICCV)},
  volume = {},
  number = {},
  pages = {17464-17473}
}

@inproceedings{ronneberger2015unet,
 author = {Ronneberger, Olaf and Fischer, Philipp and Brox, Thomas},
 booktitle = {International Conference on Medical Image Computing and Computer Assisted Intervention~(MICCAI)},
 doi = {10.1007/978-3-319-24574-4_28},
 isbn = {978-3-319-24574-4},
 pages = {234-241},
 title = {{U-Net}: Convolutional Networks for Biomedical Image Segmentation},
 year = {2015}
}

@inproceedings{loshchilov2019decoupled_adamw,
  title={Decoupled Weight Decay Regularization},
  author={Loshchilov, Ilya and Hutter, Frank},
  booktitle={International Conference on Learning Representations (ICLR)},
  year={2019},
  pages={1-19}
}

@article{olpadkar2021center,
  title={Center loss regularization for continual learning},
  author={Olpadkar, Kaustubh and Gavas, Ekta},
  journal = {arXiv preprint arXiv:2110.11314},
  year={2021}
}

@inproceedings{shrivastava2016training,
  title = {Training Region-Based Object Detectors with Online Hard Example Mining},
  author = {Shrivastava, Abhinav and Gupta, Abhinav and Girshick, Ross},
  year = {2016},
  booktitle = {IEEE Conference on Computer Vision and Pattern Recognition (CVPR)},
  volume = {},
  number = {},
  pages = {761-769},
  doi = {10.1109/CVPR.2016.89}
}

@article{lima2026toward,
  title = {Toward Unified Fine-Grained Vehicle Classification and Automatic License Plate Recognition},
  author = {G. E. {Lima} and V. {Nascimento} and E. {Santos} and E. {Nascimento Jr.} and R. {Laroca} and D. {Menotti}},
  year = {2026},
  journal = {Journal of the Brazilian Computer Society},
  volume = {32},
  number = {1},
  pages = {783-799},
  doi = {10.5753/jbcs.2026.5899},
  issn = {},
}

@inproceedings{jaderberg2015spatial,
 author = {M. {Jaderberg} and Simonyan, K. and Zisserman, A. and Kavukcuoglu, K.},
 booktitle = {International Conference on Neural Information Processing Systems~(NeurIPS)},
 pages = {2017-2025},
 title = {Spatial Transformer Networks},
 year = {2015}
}

@inproceedings{hu2018squeeze,
  title = {Squeeze-and-Excitation Networks},
  author = {Hu, Jie and Shen, Li and Sun, Gang},
  year = {2018},
  booktitle = {IEEE/CVF Conference on Computer Vision and Pattern Recognition},
  volume = {},
  number = {},
  pages = {7132-7141},
  doi = {10.1109/CVPR.2018.00745}
}

@inproceedings{he2019bag,
  title = {Bag of Tricks for Image Classification with Convolutional Neural Networks},
  author = {He, Tong and Zhang, Zhi and Zhang, Hang and Zhang, Zhongyue and Xie, Junyuan and Li, Mu},
  year = {2019},
  booktitle = {IEEE/CVF Conference on Computer Vision and Pattern Recognition (CVPR)},
  volume = {},
  number = {},
  pages = {558-567},
  doi = {10.1109/CVPR.2019.00065}
}

@inproceedings{graves2006connectionist,
 author = {Alex {Graves} and Fern\'{a}ndez, Santiago and Gomez, Faustino and Schmidhuber, J\"{u}rgen},
 booktitle = {International Conference on Machine Learning (ICML)},
 doi = {10.1145/1143844.1143891},
 pages = {369--376},
 title = {Connectionist Temporal Classification: Labelling Unsegmented Sequence Data with Recurrent Neural Networks},
 year = {2006}
}

@inproceedings{vaswani2017attention,
 author = {Ashish {Vaswani} and Noam Shazeer and Niki Parmar and Jakob Uszkoreit and Llion Jones and Aidan N. Gomez and Lukasz Kaiser and Illia Polosukhin},
 booktitle = {International Conference on Neural Information Processing Systems (NeurIPS)},
 isbn = {9781510860964},
 pages = {6000–6010},
 title = {Attention is All you Need},
 year = {2017}
}

@inproceedings{du2026mdiff4str,
  title = {{MDiff4STR}: Mask Diffusion Model for Scene Text Recognition},
  author = {Du, Yongkun and Zhao, Miaomiao and Fan, Songlin and Chen, Zhineng and Jia, Caiyan and Jiang, Yu-Gang},
  year = {2026},
  month = {Mar},
  booktitle = {AAAI Conference on Artificial Intelligence},
  pages = {3705-3713},
  doi = {10.1609/aaai.v40i5.37370}
}

@article{shi2017endtoend,
 author = {B. Shi and X. Bai and C. Yao},
 doi = {10.1109/TPAMI.2016.2646371},
 issn = {0162-8828},
 journal = {IEEE Transactions on Pattern Analysis and Machine Intelligence},
 month = {Nov},
 number = {11},
 pages = {2298-2304},
 title = {An End-to-End Trainable Neural Network for Image-Based Sequence Recognition and Its Application to Scene Text Recognition},
 volume = {39},
 year = {2017}
}

@inproceedings{woo2023convnextv2,
  title = {{ConvNeXt V2}: Co-designing and Scaling ConvNets with Masked Autoencoders},
  author = {Woo, Sanghyun and Debnath, Shoubhik and Hu, Ronghang and Chen, Xinlei and Liu, Zhuang and Kweon, In So and Xie, Saining},
  year = {2023},
  booktitle = {IEEE/CVF Conference on Computer Vision and Pattern Recognition (CVPR)},
  volume = {},
  number = {},
  pages = {16133-16142},
  doi = {10.1109/CVPR52729.2023.01548}
}

@article{oquab2024dinov2,
  title={{DINOv2}: Learning Robust Visual Features without Supervision},
  author={Maxime Oquab and others},
  journal={Transactions on Machine Learning Research},
  year={2024},
  url={https://openreview.net/forum?id=a68SUt6zFt},
  issn={2835-8856},
}

@article{simeoni2025dinov3,
  title = {{DINOv3}},
  author = {Oriane Sim{\'e}oni and others},
  year = {2025},
  journal = {arXiv preprint arXiv:2508.10104},
  number = {},
  pages = {1-67}
}

@inproceedings{carion2020end,
  title = {End-to-End Object Detection with Transformers},
  author = {Carion, Nicolas and Massa, Francisco and Synnaeve, Gabriel and Usunier, Nicolas and Kirillov, Alexander and Zagoruyko, Sergey},
  year = {2020},
  booktitle = {European Conference on Computer Vision (ECCV)},
  pages = {213-229},
  doi = {10.1007/978-3-030-58452-8_13},
  isbn = {978-3-030-58452-8}
}

@inproceedings{zhu2019deformable,
  title = {Deformable ConvNets V2: More Deformable, Better Results},
  author = {Zhu, Xizhou and Hu, Han and Lin, Stephen and Dai, Jifeng},
  year = {2019},
  booktitle = {IEEE/CVF Conference on Computer Vision and Pattern Recognition (CVPR)},
  volume = {},
  number = {},
  pages = {9300-9308},
  doi = {10.1109/CVPR.2019.00953}
}

@inproceedings{micikevicius2018mixed,
  title={Mixed Precision Training},
  author={Micikevicius, Paulius and others},
  booktitle={International Conference on Learning Representations (ICLR)},
  year={2018},
  pages={1-12}
}

@inproceedings{loshchilov2017sgdr,
  title={{SGDR}: Stochastic Gradient Descent with Warm Restarts},
  author={Loshchilov, Ilya and Hutter, Frank},
  booktitle={International Conference on Learning Representations (ICLR)},
  year={2017},
  pages={1-16}
}

@article{weihong2020research,
  title = {Research on License Plate Recognition Algorithms Based on Deep Learning in Complex Environment},
  author = {W. {Weihong} and T. {Jiaoyang}},
  year = {2020},
  journal = {IEEE Access},
  volume = {8},
  number = {},
  pages = {91661-91675},
  doi = {10.1109/ACCESS.2020.2994287}
}

@article{terven2023comprehensive,
  title={A comprehensive review of {YOLO} architectures in computer vision: From {YOLOv1} to {YOLOv8} and {YOLO-NAS}},
  author={Terven, Juan and C{\'o}rdova-Esparza, Diana-Margarita and Romero-Gonz{\'a}lez, Julio-Alejandro},
  journal={Machine Learning and Knowledge Extraction},
  volume={5},
  number={4},
  pages={1680-1716},
  year={2023},
}

@article{hsu2013application,
  title = {Application-Oriented License Plate Recognition},
  author = {G. S. Hsu and J. C. Chen and Y. Z. Chung},
  year = {2013},
  month = {Feb},
  journal = {IEEE Transactions on Vehicular Technology},
  volume = {62},
  number = {2},
  pages = {552-561},
  doi = {10.1109/TVT.2012.2226218},
  issn = {0018-9545},
}

@inproceedings{laroca2023do,
  title = {Do We Train on Test Data? {T}he Impact of Near-Duplicates on License Plate Recognition},
  author = {R. {Laroca} and V. {Estevam} and A. S. {Britto Jr.} and R. {Minetto} and D. {Menotti}},
  year = {2023},
  month = {June},
  booktitle = {International Joint Conference on Neural Networks (IJCNN)},
  volume = {},
  number = {},
  pages = {1-8},
  doi = {10.1109/IJCNN54540.2023.10191584}
}

@article{wei2024efficient,
  title = {Efficient license plate recognition in unconstrained scenarios},
  author = {Chao Wei and Fei Han and Zizhu Fan and Linrui Shi and Cheng Peng},
  year = {2024},
  journal = {Journal of Visual Communication and Image Representation},
  volume = {104},
  pages = {104314},
  doi = {10.1016/j.jvcir.2024.104314},
  issn = {1047-3203}
}

@article{wojcik2026lplcv2,
  title = {{LPLCv2}: An Expanded Dataset for Fine-Grained License Plate Legibility Classification},
  author = {L. {Wojcik} and E. A. F. {Machoski} and E. {Nascimento Jr.} and R. {Laroca} and D. {Menotti}},
  year = {2026},
  journal = {International Joint Conference on Neural Networks (IJCNN)},
  pages = {1-7},
  doi = {}
}

@inproceedings{sun2019deep,
  title = {Deep High-Resolution Representation Learning for Human Pose Estimation},
  author = {Sun, Ke and Xiao, Bin and Liu, Dong and Wang, Jingdong},
  year = {2019},
  booktitle = {IEEE/CVF Conference on Computer Vision and Pattern Recognition (CVPR)},
  volume = {},
  number = {},
  pages = {5686-5696},
  doi = {10.1109/CVPR.2019.00584}
}

@inproceedings{du2022svtr,
  title = {{SVTR}: Scene Text Recognition with a Single Visual Model},
  author = {Du, Yongkun and Chen, Zhineng and Jia, Caiyan and Yin, Xiaoting and Zheng, Tianlun and Li, Chenxia and Du, Yuning and Jiang, Yu-Gang},
  year = {2022},
  booktitle = {International Joint Conference on Artificial Intelligence (IJCAI)},
  pages = {884-890},
  doi = {10.24963/ijcai.2022/124}
}

@inproceedings{liu2024irregular,
  title = {Irregular License Plate Recognition via Global Information Integration},
  author = {Yuan-Yuan {Liu} and Liu, Qi and Chen, Feng and Yin, Xu-Cheng},
  year = {2024},
  booktitle = {International Conference on Multimedia Modeling},
  pages = {325-339},
  doi = {10.1007/978-3-031-53308-2_24},
  isbn = {978-3-031-53308-2}
}
